\documentclass{article}
\usepackage{iclr2027_conference,times}
\usepackage[utf8]{inputenc}
\usepackage[T1]{fontenc}
\usepackage{amsmath,amssymb,amsthm,booktabs,graphicx,microtype,placeins,longtable}
\usepackage{url,hyperref}
\hypersetup{hidelinks}

\newcommand{\E}{\mathbb E}
\newcommand{\R}{\mathbb R}
\newcommand{\tr}{\operatorname{tr}}
\newcommand{\rank}{\operatorname{rank}}

\newtheorem{theorem}{Theorem}
\newtheorem{proposition}[theorem]{Proposition}
\newtheorem{corollary}[theorem]{Corollary}

\title{The Model Shape Behind Scaling Laws:\\A Gradient Superposition Perspective}

\author{
  Wenjie Sun$^{1}$ \quad
  Jinning Yang$^{2}$ \quad
  Shuai Zhang$^{3}$ \quad
  Mengnan Du$^{1}$\thanks{Corresponding author.} \\
  $^{1}$The Chinese University of Hong Kong, Shenzhen \\
  $^{2}$Shenzhen Institute of Advanced Technology, Chinese Academy of Sciences \\
  $^{3}$New Jersey Institute of Technology \\
  \texttt{wenjie1835@gmail.com},~\texttt{jinny-yang@outlook.com},\\
  \texttt{shuai.zhang@njit.edu},~\texttt{mengnandu@cuhk.edu.cn}
}

\iclrfinalcopy

\begin{document}
\maketitle
\begin{abstract}
Neural scaling laws guide resource allocation, but how should a fixed parameter budget be divided between depth and width? Motivated by inverse-width representation-error scaling under strong superposition, we study this allocation through the network's end-to-end response. The Neural Feature Ansatz links weight and gradient geometry, motivating the Frobenius norm $F$ of the input average gradient outer product (AGOP) as an observable of response strength and overlap. By reproducing Anthropic's toy-model double-descent experiment, we find that $F$ and test reconstruction loss follow an approximately affine relationship across the full sample-size sweep and share an interpolation peak. Through Gaussian minimum-norm regression experiments, we further show that this correlation arises from a fitted-noise amplification term shared by prediction error and $F$. By pretraining language models spanning different shapes and parameter budgets, we find a strong positive correlation between loss and $F$ among the best shapes selected by validation loss. This supports interpreting effective parameter allocation as preserving task-relevant computation while reducing the amplification of error responses. Our results identify error-response amplification as a mechanism linking gradient superposition to generalization, providing a functional account of depth-width allocation and efficient parameter utilization.
\end{abstract}
\newcommand{\BridgeOutputError}{0.054}
\newcommand{\BridgeKLError}{0.061}
\newcommand{\BridgeFisherCorr}{0.9932}
\newcommand{\BridgeCECorr}{0.9929}
\newcommand{\BridgeTraceCorr}{0.9966}

\section{Introduction}
Neural scaling laws describe how prediction loss changes with model size, training data, and computation, and have become a practical guide to allocating resources in language modeling \citep{kaplan2020,hoffmann2022}. Their success raises a complementary question about resource utilization: \emph{once a parameter budget is fixed, how should those parameters be organized between depth and width?} Existing depth-width studies show that this allocation and its training recipe matter \citep{petty2024,mcleish2025}. A parameter count can describe both a shallow, wide network and a deep, narrow one, which need not learn equally useful functions from the same amount of data. Understanding this architectural choice complements deciding how much to scale.

Superposition provides a representation-based motivation for studying model shape. When features are sparse, a network can represent more features than it has dimensions by assigning them overlapping directions, with nonlinear computation filtering the resulting interference \citep{elhage2022}. \citet{liu2025} connect strong superposition to approximately inverse-width representation-error scaling under sufficiently balanced feature frequencies. Width is therefore a meaningful resource, but parameters also fund the layers that process representations. Since a conventional Transformer's leading parameter count scales approximately as depth times squared width, increasing width at fixed size trades away depth. This motivates asking how the representation-processing balance changes with budget. We study this balance through the response geometry of the trained input-output map.

To compare architectures, we need a quantity that includes computation across layers. The tied toy autoencoder makes feature overlap visible through $W^\top W$, but this single-bottleneck object does not directly describe a deep network's input-output behavior\citep{henighan2023}. The Neural Feature Ansatz (NFA) connects layerwise weight Gram matrices to corresponding average gradient outer products (AGOPs), motivating a gradient perspective \citep{radhakrishnan2023,beaglehole2024}. We study the full input AGOP and its Frobenius norm, $F=\|G\|_{\mathrm F}$, as an observable signature of learned response strength and overlap. We use \emph{gradient superposition} for end-to-end response overlap: magnitude together with directional sharing in the input AGOP. Measured after training, $F$ is a functional observable rather than a count of semantic features.

We next ask whether this observable tracks generalization error and helps explain depth-width allocation. In the tied-autoencoder double-descent setting of \citet{henighan2023}, we find that $F$ and test reconstruction loss follow an approximately affine relation across the full sample-size sweep and share a peak. To understand how such joint variation can arise, we analyze Gaussian minimum-norm regression. Fitting noise along weakly constrained directions produces large predictor coefficients, increasing both prediction error and input-gradient magnitude. We show that risk and $F$ contain the same fitted-noise amplification term in expectation and share an interpolation singularity under nonzero noise. Motivated by this controlled mechanism, we then test whether the loss-norm association extends to architectural variation by scanning depth and width at fixed parameter budgets. Within-budget associations vary; from 10M to 50M parameters, lower-loss shapes tend to have smaller $F$, and validation-selected shapes occupy a low-$F$ region. These observations support using $F$ to interpret how parameter allocation affects the trained network's response geometry and generalization.

Our contributions are threefold:
\begin{itemize}
    \item We formulate a gradient-superposition perspective on model shape using the Frobenius norm of the input AGOP to characterize end-to-end response strength and overlap, with explicit coordinate and normalization conventions.

    \item We identify an approximately affine loss-norm relation and a shared peak across the full sparse-reconstruction double-descent trajectory. In Gaussian minimum-norm regression, we establish a shared fitted-noise amplification term in expectation and prove a common interpolation singularity under nonzero noise, providing a controlled mechanism for joint variation in error and gradient response.

    \item Across 152 language-model shapes and twelve parameter budgets, we characterize loss-norm associations across validation-selected models and within individual budgets, identifying a low-loss, low-$F$ region from 10M to 50M. Separately, a descriptive fit to the validation-loss landscape yields a preferred-depth trajectory $d_*\propto P^{0.46}$ under the studied training recipe.
\end{itemize}

\section{Gradient superposition and the working hypothesis}\label{sec:prelim}
\subsection{A network-level observable}
Let $f:\R^D\to\R^K$ be a trained predictor, and use the output-by-input Jacobian $J_f(x)=Df(x)\in\R^{K\times D}$. For a specified evaluation distribution $\mathcal D$, define
\begin{equation}\label{eq:agop}
 G=\E_{x\sim\mathcal D}[J_f(x)^\top J_f(x)],\qquad
 F=\|G\|_{\mathrm F},\qquad \tau=\tr G.
\end{equation}
The entry $G_{ij}=\E\langle\partial_i f,\partial_j f\rangle$ measures average output-response overlap between input coordinates $i$ and $j$. Averaging the operator before taking its norm retains overlap across examples as well. We use the full norm, including its diagonal, so $F$ reflects response magnitude and spectral concentration together. It is not the average squared Jacobian norm: that quantity is $\tau=\E\|J_f\|_{\mathrm F}^2$. Appendix~\ref{app:notation} collects the notation and spectral identities used in the supplementary analysis.

The toy-model connection is explicit. For $f(x)=\operatorname{ReLU}(Hx+b)$ with $H=W^\top W$, let $D_x=\operatorname{diag}\mathbf1\{Hx+b>0\}$. The selected derivative used by automatic differentiation gives
\begin{equation}\label{eq:tied}
 J_f(x)=D_xH,\qquad G=H(\E D_x)H.
\end{equation}
Thus feature-direction geometry contributes to functional gradient geometry, together with the activation pattern. For deep networks, the corresponding layerwise NFA relation depends on architecture, initialization, and training \citep{tansley2025}; Equation~\eqref{eq:agop} directly defines our end-to-end observable. At ReLU boundaries, a derivative convention alone is insufficient for a Taylor-response claim; the additional regularity conditions are given in Appendix~\ref{app:tied}.

There is a direct reason to retain $F$ alongside trace. For one Gaussian probe $z\sim\mathcal N(0,I_D)$ shared across examples, the mean linearized response energy is $A(z)=\E_x\|J_f(x)z\|^2=z^\top Gz$. Standard quadratic-form identities give
\begin{equation}\label{eq:directional}
 \E_z A(z)=\tau,\qquad \operatorname{Var}_z A(z)=2F^2.
\end{equation}
The trace describes mean response energy, whereas $F$ describes variation across Gaussian probes, whose lengths also vary. The regression analysis below separates useful signal from amplified error within this functional response.

\subsection{From the hypothesis to architectural comparisons}
The shape sweep reallocates a fixed budget across depth and width and measures \(F\) on each trained map. Architectural comparison tests whether rearranging the same budget moves loss and \(F\) together.

Write \(P\) for active parameter count, \(d\) for depth and \(w\) for residual width. With a fixed FFN ratio and small embedding overhead, \(P\approx\kappa d w^{2}\), so increasing \(d\) at fixed \(P\) requires reducing \(w\). The scans match exact counts by adjusting FFN width and head count as well (Appendix~\ref{app:notation}). Comparisons are therefore among complete configurations under one training recipe. Norm comparisons fix the input coordinates, output convention and evaluation protocol. Orthogonal changes of the input basis preserve \(F\); input rescaling and output gain do not. Language-model measurements use the same continuous token coordinates at every width and the centered, RMS-normalized logits of Section~\ref{sec:lm}. The validation loss selects shapes and the test loss evaluates them. All candidates and seeds enter the analysis.

Depth and width are the design variables; \(F\) is measured on the trained map, aims to describe the selected response geometry. If the shape in the matched \(P\) mainly changes the amplification of the error-response, the lower-loss shapes should occupy a lower-\(F\) region of the same budget. Section~\ref{sec:lm} tests this prediction.

\section{A shared amplification mechanism}\label{sec:dd}
\subsection{Prediction and sparse-reconstruction setting}
If an amplified response component contributes to both error and input gradients, changing how strongly that component is constrained should move loss and $F$ together. Double descent provides a controlled way to examine this prediction while keeping the architecture fixed. We use the sparse reconstruction setting of \citet{henighan2023}: a tied ReLU autoencoder has input dimension $D=1000$ and bottleneck dimension two. Features are independently active with probability $0.01$, active values are uniform on $[0,1)$, and nonzero examples are normalized to unit length. The reconstruction target is the input itself.

The sweep includes 22 training sizes from $n=3$ to $40{,}000$, five initializations at each size, 3,000 AdamW steps, and 5,000 test examples. Test loss $L$ is summed squared reconstruction error per example. Equation~\eqref{eq:tied} computes the AGOP from activation frequencies without materializing every Jacobian. We retain the full range and the same training protocol throughout. Appendix~\ref{app:protocol} gives the optimizer, dataset seeds and all recorded losses; initialization error bars do not measure uncertainty from resampling the datasets.

\begin{figure}[t]
\centering\includegraphics[width=\linewidth]{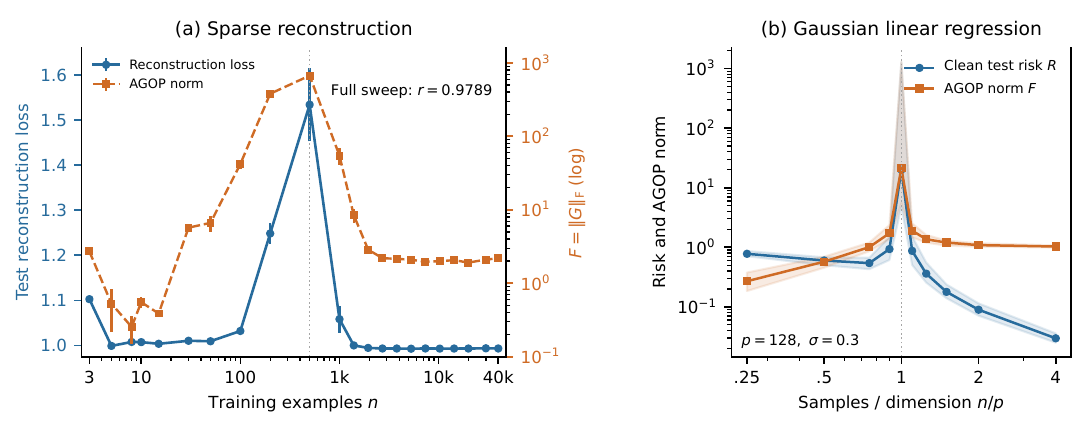}
\caption{Joint loss--norm trajectories. (a) The complete sparse-reconstruction sweep: test reconstruction loss on the left axis and AGOP norm on the logarithmic right axis, with five-initialization means and SEM at all 22 sizes. Both peaks occur at $n=500$. (b) Gaussian minimum-norm regression with $p=128$ and noise standard deviation $0.3$: clean risk and $F$, medians and 10--90\% intervals over 128 paired repetitions. The dotted line is $n=p$. The noiseless comparison and matched reconstruction baselines are in the appendix.}\label{fig:dd}
\end{figure}

Across the complete trajectory, loss and $F$ have a common peak at $n=500$ and configuration-mean Pearson correlation $r=0.9789$. The first decline, subsequent rise and later decline can all be inspected in Figure~\ref{fig:dd}(a). Fitting every configuration mean gives
\begin{equation}\label{eq:dd_affine}
 L\simeq L_0+aF,\qquad L_0=1.0017,\quad a=7.60\times10^{-4},\quad R^2=0.9581.
\end{equation}
This supports an empirical proportional trend after an estimated offset. It does not make $L-L_0$ a nonnegative error component: later losses are below that intercept. The observed common peak is the phenomenon to explain. Its training loss is not zero, so it is not identified with exact interpolation merely because it resembles a double-descent peak. The next model isolates an interpolation mechanism for which the connection can be proved.

\subsection{Minimum-norm regression isolates the shared term}
Consider $y=X\beta_*+\varepsilon$, where $X\in\R^{n\times p}$ has independent standard Gaussian entries and $\varepsilon\sim\mathcal N(0,\sigma^2I_n)$ is independent training noise. Fit $f(x)=x^\top\widehat\beta$ with $\widehat\beta=X^\dagger y$. Independent test inputs have second moment $I_p$. Throughout this section, $R$ denotes clean population prediction risk, while $S=\|\beta_*\|^2$ is fixed signal energy. This notation separates population risk in the controlled model from the measured reconstruction loss $L$.

\begin{proposition}[A common fitted-noise contribution]\label{thm:linear}
For each trained predictor,
\begin{equation}\label{eq:linear_identity}
 F=\|\widehat\beta\|^2,\qquad
 R=\|\widehat\beta-\beta_*\|^2=F+S-2\widehat\beta^\top\beta_*.
\end{equation}
Let $\Pi_X=X^\dagger X$ project onto the row space of $X$, and let $s_j(X)$ be its nonzero singular values. Conditional on $X$, averaging over training noise gives
\begin{align}\label{eq:shared_noise}
 \E_\varepsilon[F\mid X]&=\|\Pi_X\beta_*\|^2+V(X),\nonumber\\
 \E_\varepsilon[R\mid X]&=\|(I-\Pi_X)\beta_*\|^2+V(X),
 \qquad V(X)=\sigma^2\sum_{j:s_j>0}s_j(X)^{-2}.
\end{align}
\end{proposition}
The risk decomposition is classical \citep{hastie2022}. Its expression through predictor AGOP reveals that the same noise energy enters error and gradient sensitivity with coefficient one. To see the source, write the fitted-noise component as $X^\dagger\varepsilon=\sum_jv_j(u_j^\top\varepsilon)/s_j$. A direction weakly constrained by training inputs has a small nonzero singular value. Fitting noise along it creates a large predictor coefficient, and that coefficient is also an input derivative. Prediction error and the norm therefore register the same amplification. Structural zero singular values are excluded by the pseudoinverse.

This reasoning gives a precise common-peak statement near interpolation. For deterministic teacher sequences with $\|\beta_*\|^2\to S<\infty$, take $n,p\to\infty$ at fixed $p/n\to\gamma\ne1$ before approaching one. Let $\mathcal R$ and $\mathcal F$ denote the limiting risk and norm. Their shared Gaussian inverse-spectrum term yields the following consequence.
\begin{corollary}[Shared interpolation singularity]\label{cor:sharedpeak}
For $\sigma>0$,
\begin{equation}\label{eq:sharedpole}
 \lim_{\gamma\to1^\pm}|\gamma-1|\mathcal F(\gamma)
 =\lim_{\gamma\to1^\pm}|\gamma-1|\mathcal R(\gamma)=\sigma^2,
 \qquad \lim_{\gamma\to1^\pm}\mathcal R(\gamma)/\mathcal F(\gamma)=1.
\end{equation}
\end{corollary}
Appendix~\ref{app:linear} proves the conditional identities and finite-dimensional expectations; Appendix~\ref{app:sharedpeak} supplies convergence assumptions and a sufficient finite-scan peak criterion. For a realized predictor, $|R-F|\leq S+2\sqrt{SF}$ also shows why bounded signal and target alignment cannot dominate when $F$ becomes large. The contribution is the joint AGOP--risk interpretation of this classical instability, rather than a new derivation of ridgeless risk in isolation.

\subsection{Testing the mechanism and interpreting both descents}
Figure~\ref{fig:dd}(b) tests this mechanism with $p=128$, $\beta_*=e_1$, ten sample sizes and 128 paired repetitions. At noise standard deviation $0.3$, risk and norm rise together near $n=p$ and decline as the inverse-spectrum amplification weakens. Medians and 10--90\% intervals describe the heavy-tailed interpolation region. The companion noiseless result is informative: for $n<p$, $R=S-F$, and the noise-driven singularity is absent (Appendix~\ref{app:linear}). Thus the theorem specifies the component responsible for a positive relation; it does not assert that every input-dependent signal raises error.

Target alignment can also be controlled exactly. Holding a unit teacher fixed and fitting only noise in orthogonal directions gives $R=F-1$ for every realization; for a zero teacher, $R=F$. Sixty-four paired simulations verify this construction (Appendix~\ref{app:controlledsignal}). It identifies the target-alignment condition under which proportionality holds exactly.

These results suggest a gradient interpretation of the two descents. With few observations, fitting individual patterns need not create a large response on most unseen inputs. Near a poorly constrained regime, larger response coefficients can increase both norm and prediction error. Additional observations can then reduce that amplification. In the linear model, these statements follow from the nonzero singular spectrum; in the sparse autoencoder, the joint trajectory is an observed association. The latter has noiseless targets, learned gates and biases, so the Gaussian mechanism is a controlled explanation of how a shared peak can arise, not an identified noise source for the toy task.

The interpretation also distinguishes reduced amplification from successful feature recovery. In the late toy regime, the matched training-mean predictor achieves loss $0.992437$ at $n=40{,}000$, versus $0.993247$ for the network, while having zero AGOP. The later decline of $F$ therefore accompanies lower error but does not by itself establish input-dependent shared-feature learning. Appendix~\ref{app:toycontrols} reports the measured reconstruction decomposition and matched baseline comparison. The central result remains the full-trajectory association, together with a regression model in which joint amplification is mathematically explicit.

\section{Model shape under a fixed parameter budget}\label{sec:lm}
\subsection{Architectural prediction and scan protocol}
The amplification account predicts that, within a parameter budget, shapes with lower generalization error should also occupy a region of smaller functional sensitivity. We test the loss--norm relationship both across validation-selected shapes at different budgets and among shapes within each budget.

The scan comprises 254 byte-level Transformer checkpoints and 152 configuration means across twelve budgets from 0.3M to 50M on FineWeb-derived data \citep{penedo2024}. Depth and width vary within each budget; FFN width is adjusted to match active parameters within 0.5\%. Vocabulary size and context length are both 256, and attention-head dimension is four. Head count and FFN ratio therefore change with residual width. Available initialization seeds are averaged within each configuration. Every candidate's exact parameters, exposure, dimensions, attention heads and validation/test losses are listed in Appendix~\ref{app:lmtables}. The seven 15M shapes enter the same configuration-mean landscape, frontier and allocation fit as the other budgets; 15M is not reserved for independent shape validation.

Data exposure increases with model size, following the allocation motivation of Chinchilla \citep{hoffmann2022}. The recorded parameter--data schedules span roughly 96--306 processed bytes per parameter. Pooled relationships describe this training path, while within-budget comparisons hold nominal model size fixed. Increasing depth can alter optimization and inference cost even at constant parameter count. The corpus, optimizer and checkpoint-selection details are documented in Appendix~\ref{app:protocol} and Table~\ref{tab:allshapes}.

\subsection{A common measurement space}
To compare widths without changing the differentiation space, extend tokens to continuous one-hot coordinates $q\in\R^{C\times V}$ and embed them as $e_t=q_tE+p_t$. Here $C=V=256$, so the input dimension is always $CV=65{,}536$. Let $\ell(q)$ be the last-position logits, $C_V=I-\mathbf1\mathbf1^\top/V$, and $s_\ell>0$ their centered RMS on the probe set. The function used in Equation~\eqref{eq:agop} is
\begin{equation}\label{eq:lm_function}
 f(q)=\frac{C_V\ell(q)}{s_\ell},\qquad
 s_\ell^2=\E_{\rm probe}\|C_V\ell(q)\|^2/V.
\end{equation}
Centering removes a common logit shift. RMS normalization reduces the direct contribution of output gain, with $s_\ell$ held constant during differentiation. The embedding remains inside the function, so the measurement is end to end. We use 96 examples and 96 shared probes, averaging operator products across examples before taking their squared norm. Cross-entropy is evaluated with raw logits and averaged across token positions. Thus AGOP and CE have specified, different output aggregations; neither their units nor their numerical magnitudes are equated.

The resulting quantity depends on the selected continuous extension of discrete tokens. Appendix~\ref{app:estimator} gives the estimator, and Appendix~\ref{app:newaudit} reports raw-logit, tangent-coordinate, output-position and sample/probe controls. All main-scan associations use this common measurement convention; the controls quantify its coordinate and sampling dependence.

\subsection{Across-budget and within-budget associations}
\begin{figure}[t]
\centering\includegraphics[width=\linewidth]{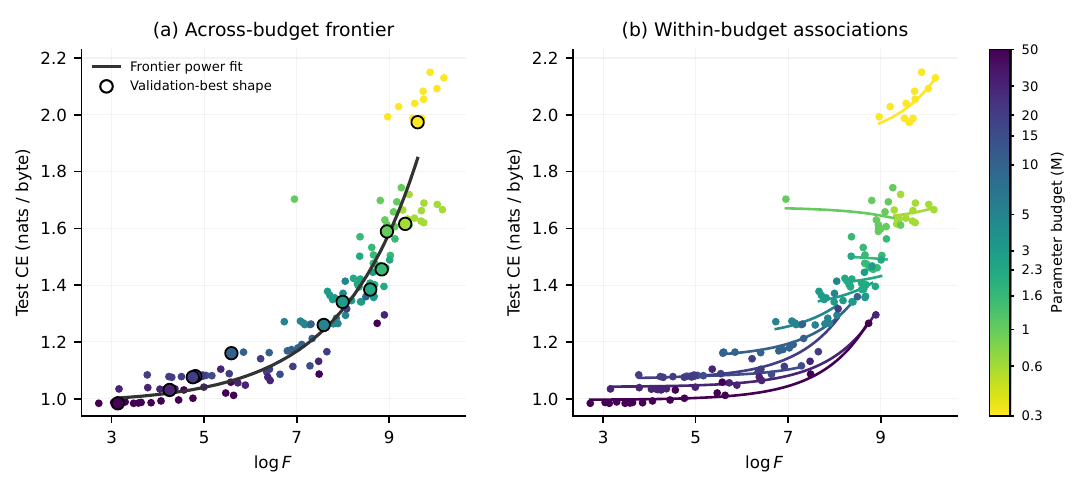}
\caption{Loss--norm associations at two allocation scales. Both panels show all 152 configuration means against $\log F$ and share a continuous color scale: larger parameter budgets are darker. (a) Black outlines mark one validation-best shape per budget; the curve fits the test loss of these twelve frontier points. (b) Each budget has an unconstrained least-squares fit $L_{\mathrm{test}}=aF+bF^2$, displayed on the same logarithmic-norm coordinate. Lines cover observed norm ranges only; they are descriptive fits, not paths through untrained architectures.}\label{fig:lm}
\end{figure}

We first select the shape with the smallest seed-mean selection-validation CE at each budget. These twelve points define a descriptive frontier. Their test CE and $F$ have Pearson $r=0.9733$ and Spearman $\rho=1.000$. Figure~\ref{fig:lm}(a) summarizes the trend with
\begin{equation}\label{eq:frontierfit}
 \widehat L_{\rm test}(F)=0.9839+0.1734(F/1000)^{0.5929},
 \qquad R^2=0.9448.
\end{equation}
Selection uses validation loss before this test-loss fit. The fit therefore describes the norm accompanying validation-based architectural choices; it does not select them from test labels. Its intercept reaches the fitting bound and is not an identified irreducible loss. The smooth curve provides a compact view of a finite frontier rather than a universal scaling law in $F$.

The frontier combines changes in budget with changes in preferred architecture. Figure~\ref{fig:lm}(b) tests the prediction at fixed budget, fitting every candidate and allowing either sign of slope. The fit is linear in $F$ and therefore appears curved against $\log F$. Across all 152 configuration means, the unlogged-$F$/test-loss correlation is $0.8801$. Within budgets, it is at least $0.95$ from 10M onward, supporting the amplification account when the same parameter budget is rearranged. The pairing is regime-dependent: the 1M and 1.6M budgets have $r=-0.188$ and $-0.056$, respectively. These weak or negative associations mark the small-budget boundary of the observed joint response. Appendix~\ref{app:correlations} reports every budget and its depth-dependent curves.

The two scales address distinct allocation questions: the frontier tracks favorable shapes as resources grow, while within-budget curves track how those resources are rearranged. Their agreement at larger budgets shows that the loss--norm pairing persists after conditioning on nominal parameter count, so variation in budget alone cannot explain it.

\subsection{How the favorable shape changes with budget}\label{sec:shapefit}
The allocation question also concerns the shape itself. Among the scanned candidates, the validation-selected depth rises from six at 0.3M to 20 at 10M, 24 at 15M and 20M, 32 at 30M, and 40 at 50M. Intermediate small budgets are not monotone. The broad progression motivates a descriptive fit to the full landscape, rather than an extrapolation from the best point of a single budget.

We model validation CE using a budget-dependent baseline plus a quadratic penalty in log depth. The baseline is a power law in nominal budget; the penalty's center is allowed to move linearly with log budget. This gives a continuous preferred depth while leaving the empirical candidate set discrete. All twelve budgets, including 15M, receive equal total fitting weight, and only selection-validation CE enters the fit. The main coefficients use the full landscape; leave-one-budget-out cross-validation assesses fit generalization. Appendix~\ref{app:fits} gives the expression, bounds, coefficients and all twelve folds.

The fitted center is $d_*(P)=5.296(P/10^6)^{0.459}$. Under the approximation $P\propto dw^2$, this corresponds to $w_*\propto P^{0.2705}$. The fitted center is the shape trajectory implied by the low-loss band under this recipe: both depth and width increase, with depth growing faster. The exponents describe the measured range and its parameter--data path. They do not identify separate processing and representation losses; actual widths are additionally affected by changing FFN ratios and embedding overhead.

Holding out whole budgets gives validation RMSE $0.0701$, compared with $0.0844$ for a budget-only model. A separable depth-width model gives $0.0720$ and a smaller mean validation selection gap ($0.00159$ versus $0.00605$ CE). Better prediction of the full landscape thus need not yield better selection among its best candidates. Likewise, validation loss on the frontier fits budget alone better than it fits $F$ ($R^2=0.9877$ versus $0.9374$). AGOP contributes a functional interpretation of the shape landscape, while these comparisons quantify predictive fit and candidate-selection accuracy.

\subsection{What the association says about allocation}\label{sec:allocation_interpretation}
We read the low-loss, low-$F$ band at larger budgets as a shape-dependent change in error-response amplification. The regression identity shows why weakly constrained directions raise risk and $F$ together when they amplify fitted error. Reallocating depth and width changes which responses the trained map amplifies. Under this account, a shape that generalizes better at matched parameter count is less sensitive in the AGOP sense. The large-budget sweeps exhibit this joint response.

Appendix~\ref{app:cleanbridge} gives a sufficient clean-CE analogue: for a shared task-logit function and a common error direction, CE and raw-logit AGOP grow together when the gradient cross term is controlled. Extending that prediction to the normalized $F$ requires control of output scale as well. The scan measures the predicted pairing; the latent decomposition is the working interpretation specified in Claim scope.

\paragraph{Predictions tested by the scan.}
Across budgets, validation-selected shapes should trace a joint loss--norm trajectory; the twelve-point frontier has $\rho=1$. Within the large-budget regime, rearranging a fixed $P$ should move loss and $F$ together and place validation-best shapes toward the low-$F$ part of each sweep; all five budgets from 10M to 50M have $r\geq0.95$. Persistence within a budget rules out variation in nominal parameter count as the sole explanation. The weak or negative small-budget associations establish that $F$ is not simply a relabeling of loss.

\paragraph{Tests that could overturn the account.}
A systematic decrease in clean CE accompanied by increasing $F$ under compute-matched shape interventions throughout the large-budget regime would overturn this architectural prediction. A direct mechanism test would intervene on input-dependent error responses at fixed architecture and track task alignment, cross-gradient terms and output scale. Repeated improvements in clean CE without reduced scale-controlled AGOP, after ruling out pure class-bias shifts, would argue against error-response amplification as the dominant explanation of the band. These are prospective tests, distinct from the descriptive fits evaluated here.

The supplementary local-perturbation measurements in Appendix~\ref{app:perturbation} use frozen checkpoints already included in the fitting data. They check input sensitivity, without selecting frontier points or validating the allocation fit. Local CE curvature can reverse an $F$ ordering because it also depends on probability weighting and nonlinear logit curvature. The amplification account therefore concerns the observed shape region under the stated measurement convention. It connects the allocation of representation width and processing depth to the error responses of the trained function.

\section{Related work}
\subsection{Scaling laws}\label{sec:rw_scaling}
Early measurements found predictable learning curves across tasks \citep{hestness2017}, followed by joint model-data error models that extrapolated across scales \citep{rosenfeld2020}. \citet{kaplan2020} quantified power-law dependence on parameters, data and training compute for language models; \citet{henighan2020} extended this pattern to other autoregressive domains. Increasing scale also improved measured few-shot capabilities \citep{brown2020}, making scaling a practical design strategy. The question then shifted from adding resources to allocating them: Chinchilla revised the compute-optimal balance between model size and training tokens \citep{hoffmann2022}. Data pruning \citep{sorscher2022} and data-constrained scaling \citep{muennighoff2025} further showed that the value of additional resources depends on how examples are selected or reused. Theoretical accounts distinguish variance- and resolution-limited regimes \citep{bahri2024}, while Gemstones demonstrates the dependence of fitted prescriptions on architecture and training choices \citep{mcleish2025}. These developments motivate a complementary efficiency question: after fixing a parameter budget and data schedule, is that capacity organized effectively? Parameter count alone leaves open how much is assigned to representation width versus processing depth.

\subsection{Model shape}\label{sec:rw_shape}
Shape studies approach this allocation from different objectives. Depth-separation theory establishes expressive advantages of deep networks \citep{telgarsky2016}; EfficientNet balances depth, width and resolution for empirical vision efficiency \citep{tan2019}. For Transformers, \citet{levine2020} derive a width-dependent transition in self-attention depth efficiency, whereas \citet{petty2024} compare fixed-parameter shapes through language modeling and compositional generalization, finding diminishing depth returns. Gemstones studies architecture-dependent scaling \citep{mcleish2025}; CompleteP shows that depth-aware parameterization and optimization can change the efficiency of deep models \citep{dey2025}. \citet{liao2026} additionally vary residual width and internal FFN/attention bottlenecks. Mechanistic accounts relate inverse-depth behavior to similarity across layers \citep{liu2026depth} and preferred shapes to scaling coefficients \citep{liu2026coefficients}. These perspectives address expressivity, task performance or training efficiency, with conclusions tied to their model families and recipes. A complementary representation argument links strong superposition to approximately inverse-width error scaling \citep{liu2025}. That width law alone leaves processing depth unspecified: at fixed parameters, widening also removes layers. This motivates our study of how the depth-width balance appears in trained input-output response geometry, alongside validation performance, rather than prescribing shape from width alone.

\subsection{Superposition hypothesis}\label{sec:rw_superposition}
The superposition hypothesis was originally introduced to explain the phenomenon of neuron polysemanticity \citep{olah2020}, where a single neuron usually activates to multiple different concepts. This hypothesis posits that polysemanticity arises from the limited width of neural networks, which forces different features to be encoded in non-orthogonal directions within the activation space. \citet{elhage2022} examined this geometry through the encoder--decoder weight Gram matrix in a tied toy autoencoder. This work also catalyzed subsequent research aimed at resolving polysemanticity through the use of Sparse Autoencoders (SAEs) \citep{gao2025,rajamanoharan2024,bricken2023,huben2024,sun2026price}.
These methods help disentangle polysemantic representations without removing superposition from the original model. The classical $W^\top W$ analysis nevertheless concerns a tied single bottleneck, and many SAEs analyze one activation space. Crosscoders already extend feature extraction across layers \citep{lindsey2024}, but an activation dictionary does not directly characterize a deep model's complete input-output sensitivity. NFA offers a complementary connection between layerwise weight Gram matrices and corresponding AGOPs \citep{radhakrishnan2023}; alignment dynamics \citep{beaglehole2024} and depth-dependent theory \citep{tansley2025} clarify its conditions. Motivated by this connection, we study end-to-end response overlap through AGOP and its Frobenius norm. This gradient-superposition perspective characterizes functional response sharing across the full computation, complementing feature dictionaries in activation space. Appendix~\ref{app:related} gives further comparisons.

\section{Discussion and limitation}

\paragraph{When does $F$ track generalization?}
Our analysis identifies a mechanism-dependent relationship between gradient response and generalization. In minimum-norm regression, recovered signal contributes positively to F while reducing prediction risk, whereas amplified fitted noise increases both. Their joint variation is therefore determined by the balance between changes in signal recovery and error amplification. The noiseless reversal provides an exact illustration of this distinction. In the language-model scans, the strong within-budget associations across all five budgets from 10M to 50M establish that the observed pairing persists when nominal parameter count is held fixed. These results support our interpretation of the low-loss, low-F region through reduced error-response amplification. The corresponding decomposition remains to be identified directly in language models. Operationally, F is most informative for comparisons made under a common task and measurement convention, with validation performance anchoring task quality. The coordinate and sampling controls further distinguish broad loss–norm associations from fine ordering among nearby shapes. This perspective gives the observable a concrete role: characterizing how architectural allocation changes the response geometry accompanying generalization.

\paragraph{Effective depth and looped Transformers.}
The amplification account makes a further prediction for effective depth: when additional sequential processing improves clean CE at fixed stored parameters, the trained map should also have smaller $F$. Universal Transformers share parameters across depth \citep{dehghani2019}, and recurrent-depth language models use extra latent computation at inference \citep{geiping2025}. Looping changes compute as well as effective depth, and iso-depth recurrence studies show that repeated and unique blocks can differ \citep{schwethelm2026}. Comparing loop count and unique depth at matched stored parameters, exposure and inference compute would test whether useful sequential computation reduces error amplification, or whether unstable Jacobian products reverse the predicted pairing.

\paragraph{Scope of the evidence.}
The scan uses byte-level models, one source corpus and at most 50M parameters. Parameter matching leaves compute and depth-specific tuning unmatched, so we do not claim a training rule for large-scale pretraining. The amplification mechanism is identified in regression and is our interpretation of the language-model association. Establishing it as the cause of shape preference requires an intervention that changes error amplification at fixed architecture, or changes architecture at matched compute.

\section*{AI use statement}

In this work, we used generative AI tools to implement methods. We have not used generative AI tools to generate synthetic datasets, develop theoretical models, formulate or prove mathematical claims, propose hypotheses, design experiments, assist with translation, clean datasets, perform qualitative analysis, or interpret results; the remaining required-disclosure tasks are not applicable. Additionally, we used generative AI tools to polish writing, browse and summarize literature, and format references. We reviewed all AI-assisted code and text, verified citations against original sources, and take responsibility for the final content of this work.

\section*{Reproducibility statement}
The source package retains the measurements and supplementary controls, with scripts for the revised figures. The appendix supplies proofs and protocols. Appendix~\ref{app:lmtables} lists all twelve budgets used in the frontier and allocation fit; the underlying per-seed measurements are retained in the data archive.

\clearpage
\bibliography{iclr2027_conference}
\bibliographystyle{iclr2027_conference}
\clearpage
\appendix
\numberwithin{equation}{section}
\section{Notation and supplementary definitions}\label{app:notation}
The main text uses $L$ for measured reconstruction loss or cross-entropy, with the task specified, and $R$ for clean population risk in the linear control. The input AGOP is always $G=\E[J_f^\top J_f]$; $F$ denotes its full Frobenius norm, not its squared norm or trace. Additional symbols below are local to the indicated setting.
\begin{table}[htbp]\centering\small
\begin{tabular}{ll}\toprule
Symbol & Meaning \\ \midrule
$f,J_f,G$ & Predictor, output-by-input Jacobian, input AGOP \\
$F,\tau$ & $\|G\|_{\mathrm F}$ and $\tr G$ \\
$L,R$ & Task loss; clean population risk in linear regression \\
$P_0,P,d,w$ & Nominal budget, active parameters, depth, residual width \\
$m,h,a$ & FFN width, head count, head dimension in model tables \\
$n,p,\beta_*,S$ & Regression sample count, dimension, teacher, $\|\beta_*\|^2$ \\
$\sigma^2,V(X)$ & Training-noise variance and amplified fitted-noise energy \\
$H,D_x,b$ & Toy weight Gram matrix, activation gate, decoder bias \\
$q,\ell,s_\ell$ & Continuous token coordinates, raw logits, centered-logit RMS \\
$r_{\rm eff}$ & Effective spectral rank of AGOP \\
$K,\Gamma$ & CE Fisher operator and nonlinear logit-curvature contribution \\ \bottomrule
\end{tabular}
\caption{Notation. The teacher $\beta_*$ and toy bias $b$ are distinct. Attention-head dimension $a$ is used only in architecture tables; error-amplitude parameters are defined separately in Appendix~\ref{app:cleanbridge}.}
\end{table}

\paragraph{Exact model size.}
For our bias-free projections, vocabulary $V$, context length $C$, and untied output head,
\begin{equation}\label{eq:budget}
 P=w\big[d(4w+2m+4)+(2V+C+2)\big],\qquad a=w/h.
\end{equation}
The $P\approx\kappa dw^2$ approximation additionally fixes FFN ratio and neglects smaller embedding terms. All reported parameter matching uses the exact count. With candidate set $\mathcal A(P_0)$ and seed-mean selection-validation loss $\bar L_{\rm val}$, define
\begin{equation}\label{eq:validation_minimum}
 L_*(P_0)=\min_{s\in\mathcal A(P_0)}\bar L_{\rm val}(s).
\end{equation}
Here $s=(d,w,m,h)$. Frontier points use this minimizer at each budget. The 15M budget uses the same rule.

\paragraph{Spectrum and normalization.}
For nonzero positive semidefinite $G$ with eigenvalues $\lambda_i$,
\begin{equation}\label{eq:spectrum}
 F^2=\sum_{i,j}G_{ij}^2=\sum_i\lambda_i^2,\qquad
 r_{\rm eff}=\frac{\tau^2}{F^2},\qquad \tau=\sqrt{r_{\rm eff}}F.
\end{equation}
Thus $1\leq r_{\rm eff}\leq\rank G$. Effective rank is not a semantic feature count. A fixed output gain $c$ multiplies $G$ and $F$ by $c^2$. For centered language-model logits, the unnormalized operator $\widetilde G$ satisfies $\|\widetilde G\|_{\mathrm F}=s_\ell^2F$. An unlogged value of the normalized $F$ must therefore not be confused with an unnormalized logit norm.

\section{Linear regression: proof and interpretation}\label{app:linear}
\subsection{Predictor and residual AGOP}
Condition on a trained $\widehat\beta$. The predictor has constant Jacobian $J_f=\widehat\beta^\top$, so $G_f=\widehat\beta\widehat\beta^\top$. A rank-one positive semidefinite matrix $uu^\top$ has Frobenius norm $\|u\|^2$. With independent test inputs satisfying $\E xx^\top=I_p$, the squared prediction error gives the risk identity in Equation~\eqref{eq:linear_identity}. Cauchy--Schwarz gives
\begin{equation*}
 |R-F|\leq S+2\sqrt{SF}.
\end{equation*}
Thus $R/F\to1$ along realizations with $F\to\infty$ and bounded $S$. Away from this regime, target alignment can matter substantially. In contrast, for the residual $r=f-f_*$, $G_r=(\widehat\beta-\beta_*)(\widehat\beta-\beta_*)^\top$ and $\|G_r\|_{\mathrm F}=R$ exactly. The predictor and residual operators must not be interchanged. Independent noisy test labels add $\sigma^2$ to clean risk.

For non-isotropic test second moment $\Sigma$, clean risk becomes $R=(\widehat\beta-\beta_*)^\top\Sigma(\widehat\beta-\beta_*)$. It equals $\|\Sigma^{1/2}G_r\Sigma^{1/2}\|_{\mathrm F}$, not generally $\|G_r\|_{\mathrm F}$. For a multi-output linear residual $r(x)=\Delta^\top x$, isotropy gives $R=\tr(\Delta\Delta^\top)=\sqrt{r_{\rm eff}(G_r)}\|G_r\|_{\mathrm F}$. The exact scalar norm identity uses rank one.

\subsection{Training-averaged noise term}
When $n>p+1$, Gaussian $X$ has full column rank almost surely, and
\begin{equation*}
 \widehat\beta=\beta_*+\eta,\qquad
 \eta=(X^\top X)^{-1}X^\top\varepsilon,\qquad
 \E[\eta\eta^\top\mid X]=\sigma^2(X^\top X)^{-1}.
\end{equation*}
The cross term has conditional mean zero. Therefore $\E[F\mid X]=S+V(X)$ and $\E[R\mid X]=V(X)$. The inverse-Wishart expectation gives $\E V(X)=\sigma^2p/(n-p-1)$.

When $p>n+1$, let $\Pi=X^\top(XX^\top)^{-1}X$ be the orthogonal row-space projector. Then
\begin{equation*}
 \widehat\beta=\Pi\beta_*+\eta,\qquad
 \eta=X^\top(XX^\top)^{-1}\varepsilon.
\end{equation*}
The noise vector lies in the row space, while $(I-\Pi)\beta_*$ lies in its orthogonal complement. Hence
\begin{equation*}
 R=\|(I-\Pi)\beta_*\|^2+\|\eta\|^2.
\end{equation*}
Conditional mean zero again removes the signal--noise cross term from $\E F$. Rotational invariance gives $\E\Pi=(n/p)I_p$, and the inverse-Wishart formula gives $\E\|\eta\|^2=\sigma^2n/(p-n-1)$. Combining the two regimes gives the classical finite-dimensional expectations \citep{hastie2022}:
\begin{align}\label{eq:linear_mean}
 \E F&=S+\E V,&\E R&=\E V &&(n>p+1),\nonumber\\
 \E F&=\tfrac np S+\E V,&\E R&=(1-\tfrac np)S+\E V &&(p>n+1),
\end{align}

For the singular value decomposition restricted to nonzero singular values,
\begin{equation*}
 \eta=\sum_jv_j\frac{u_j^\top\varepsilon}{s_j},\qquad
 \E[\|\eta\|^2\mid X]=\sigma^2\sum_js_j^{-2}=V(X).
\end{equation*}
The design matrix $X$ is the specific ``fitting matrix'' in this argument. Structural zero eigenvalues in an overparameterized Gram matrix are excluded by the pseudoinverse; amplification comes from small \emph{nonzero} singular values. At $n=p$ and adjacent dimensions, the relevant inverse moments have infinite expectations for nonzero noise, although fitted realizations are almost surely finite. In the proportional limit, first letting $n,p\to\infty$ with $p/n\to\gamma\ne1$, the noise contribution is $\sigma^2\gamma/(1-\gamma)$ for $\gamma<1$ and $\sigma^2/(\gamma-1)$ for $\gamma>1$. It diverges as $\gamma\to1$. Without noise there is no such divergent term.

\subsection{Numerical control}
\begin{figure}[htbp]\centering\includegraphics[width=.85\linewidth]{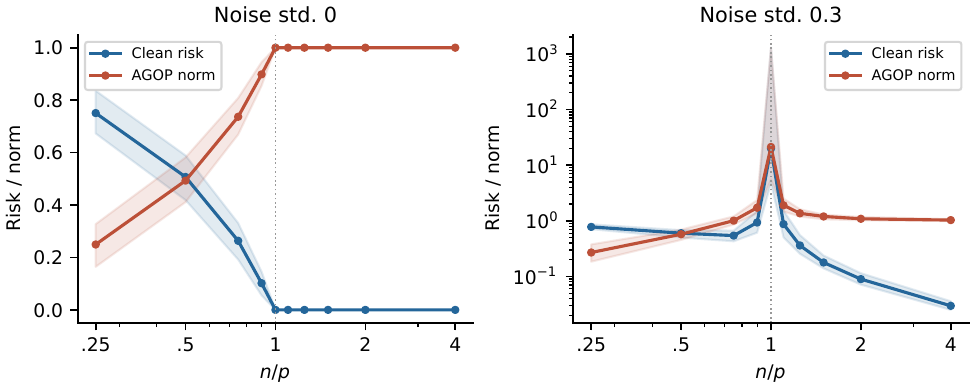}
\caption{Gaussian minimum-norm regression: the complete sample-size scan for noise standard deviations 0 and 0.3. Medians and 10--90\% intervals over 128 paired repetitions. Only the noisy case has an interpolation singularity; in the noiseless underdetermined regime $R=S-F$.}\label{fig:gaussian}
\end{figure}

The saved experiment uses $p=128$, $\beta_*=e_1$, sample sizes $n\in\{32,64,96,115,128,\allowbreak141,160,192,256,512\}$, and noise standard deviations $\{0,0.1,0.3\}$. Each of 128 repetitions generates a $512\times128$ Gaussian matrix and a length-512 noise vector; their prefixes pair sample sizes. An SVD solution is computed with relative cutoffs $10^{-12}$ and $10^{-10}$. Figure~\ref{fig:gaussian} uses cutoff $10^{-12}$. Heavy tails near interpolation motivate medians and 10--90\% intervals rather than an unstable Monte Carlo mean. Table~\ref{tab:linearcontrol} compares the noiseless and noisy cases. Identity checks use per-run values, not separately aggregated medians.
\begin{table}[htbp]
\centering\small\begin{tabular}{rrrrr}\toprule
$n$ & $R_0$ & $F_0$ & $R_{.3}$ & $F_{.3}$ \\ \midrule
32 & 0.75061 & 0.24939 & 0.77733 & 0.26958 \\
64 & 0.50695 & 0.49305 & 0.60008 & 0.57358 \\
96 & 0.26351 & 0.73649 & 0.54243 & 1.0069 \\
115 & 0.10173 & 0.89827 & 0.93252 & 1.7179 \\
128 & 5.9547e-29 & 1 & 19.922 & 21.197 \\
141 & 6.5099e-30 & 1 & 0.87065 & 1.8991 \\
160 & 4.4096e-30 & 1 & 0.36091 & 1.3644 \\
192 & 3.7739e-30 & 1 & 0.17854 & 1.1978 \\
256 & 3.1292e-30 & 1 & 0.089698 & 1.0875 \\
512 & 3.0632e-30 & 1 & 0.029712 & 1.0299 \\
\bottomrule\end{tabular}

\caption{Gaussian control, medians over 128 paired repetitions. $R_0,F_0$ use $\sigma=0$; $R_{.3},F_{.3}$ use $\sigma=0.3$. All use $p=128$ and cutoff $10^{-12}$. The noisy interpolation peak is absent in the noiseless control.}\label{tab:linearcontrol}
\end{table}

\subsection{Proof of the common interpolation singularity}\label{app:sharedpeak}
For fixed signal energy $S$ and noise variance $\sigma^2>0$, Proposition~\ref{thm:linear} gives the Gaussian proportional-limit expressions \citep{hastie2022}
\begin{align*}
 \mathcal F(\gamma)&=S+\frac{\sigma^2\gamma}{1-\gamma},&
 \mathcal R(\gamma)&=\frac{\sigma^2\gamma}{1-\gamma},&&0<\gamma<1,\\
 \mathcal F(\gamma)&=\frac S\gamma+\frac{\sigma^2}{\gamma-1},&
 \mathcal R(\gamma)&=\left(1-\frac1\gamma\right)S+\frac{\sigma^2}{\gamma-1},&&\gamma>1.
\end{align*}
Precisely, take a deterministic sequence $\beta_{*,p}$ with $\|\beta_{*,p}\|^2\to S<\infty$, independent standard Gaussian design entries, and independent Gaussian noise with fixed variance $\sigma^2>0$. Let $n,p\to\infty$ with $p/n\to\gamma\in(0,1)\cup(1,\infty)$. The isotropic risk limit in \citet[Theorem~1]{hastie2022} applies. Concentration of the Gaussian noise quadratic form and of the random row-space projection also gives $F\to\mathcal F(\gamma)$ in probability; the signal--noise inner product vanishes in probability at a fixed aspect ratio away from one. For example, away from the interpolation ratio the conditional noise variance is $2\sigma^4\tr[(X^\top X)^{-2}]=O(1/n)$ in the full-column-rank regime; the signal--noise cross variance is also $O(1/n)$. The nonzero spectrum of $XX^\top$ and concentration of the Haar row-space projection give the corresponding underdetermined result. These facts, rather than convergence of expectations alone, justify the random-quantity limits. Only after this inner limit do we let $\gamma\to1$. Multiplying by $|\gamma-1|$ leaves $\sigma^2$, while bounded signal terms vanish, and the ratio tends to one. At finite $p,n$, the square and immediately neighboring inverse-Wishart means diverge even though fitted realizations are almost surely finite.

The realization-level Cauchy--Schwarz bound in Appendix~\ref{app:linear}, divided by $F>0$, proves the second part of Corollary~\ref{cor:sharedpeak}. Along any sequence with $F\to\infty$ and bounded signal, $R/F\to1$. This reads the classical inverse-spectrum instability directly as amplification of the predictor's input-gradient operator.

\paragraph{A finite-scan peak criterion.}
For a fixed teacher and a collection of fitted linear predictors, let $j$ maximize $F_i$. The triangle and reverse-triangle inequalities give
\begin{equation*}
 (\sqrt{F_i}-\sqrt S)^2\leq R_i\leq(\sqrt{F_i}+\sqrt S)^2.
\end{equation*}
If $\sqrt{F_j}-\max_{i\ne j}\sqrt{F_i}>2\sqrt S$, then $R_j>R_i$ for all $i\ne j$, so both peaks occur at $j$. The criterion is sufficient, not necessary. It makes explicit why a dominating amplification peak can survive bounded alignment differences. The shared asymptotic singularity alone does not force every finite random realization to have exactly the same maximizing sample size.

\subsection{Exact proportionality with controlled target alignment}\label{app:controlledsignal}
Let $x\in\R^{k+1}$ have identity second moment, fix $\beta_*=e_1$, and let $U=[e_2,\ldots,e_{k+1}]$. We hold the signal predictor $x^\top\beta_*$ fixed and fit only a residual $x^\top U\widehat\theta$ to noisy training targets. With $Z=XU$, minimum-norm fitting gives $\widehat\theta=Z^+\varepsilon$ and
\begin{equation*}
 \widehat\beta=\beta_*+U\widehat\theta,\qquad
 \widehat\beta^\top\beta_*=1,\qquad
 F=1+\|\widehat\theta\|^2,\qquad L=\|\widehat\theta\|^2=F-1.
\end{equation*}
Gates are identity, bias is zero, target energy and alignment remain one, and only the noise-fitting component changes. Hence the clean population loss and norm have identical peak locations at every noise realization, independently of Gaussianity. Under Gaussian $Z$ and nonzero noise, the interpolation instability occurs near $n=k$. If the fixed teacher is zero, the same construction gives $L=F$ exactly. This is a controlled model of the mechanism, rather than a claim that nuisance terms are constant in the trained autoencoder.

Our numerical control uses $k=128$, the same ten sample sizes as the Gaussian experiment, noise standard deviations 0.1 and 0.3, and 64 paired design/noise repetitions. Independent Gaussian test inputs number 8,192 per repetition. The learner observes the noisy residual and fits its coefficients by SVD; no architecture or seed is selected using test loss. Table~\ref{tab:controlledsignal} reports all sizes for $\sigma=0.3$. The realized coefficient identity is checked separately from finite-test sampling error.
\begin{table}[htbp]\centering\small
\begin{tabular}{rrrr}\toprule
$n$ & Median $F$ & Median $L=F-1$ & Empirical test loss \\ \midrule
32 & 1.0312 & 0.031162 & 0.031338 \\
64 & 1.0893 & 0.089289 & 0.089892 \\
96 & 1.2637 & 0.2637 & 0.26388 \\
115 & 1.7087 & 0.70875 & 0.7138 \\
128 & 52.624 & 51.624 & 51.482 \\
141 & 1.7733 & 0.77331 & 0.77518 \\
160 & 1.3458 & 0.34582 & 0.34363 \\
192 & 1.1891 & 0.18915 & 0.19063 \\
256 & 1.0899 & 0.089881 & 0.090048 \\
512 & 1.0285 & 0.028534 & 0.028911 \\
\bottomrule\end{tabular}

\caption{Frozen signal and target alignment: medians over 64 repetitions. The exact population identity is $L=F-1$; the last column measures squared error on 8,192 independent test inputs.}\label{tab:controlledsignal}
\end{table}

\subsection{Localized memorization can have small population AGOP}\label{app:localized}
Consider finitely many distinct points $x_i\in\R^D$, scalar targets $y_i$, and a smooth compactly supported bump $\phi$ with $\phi(0)=1$. For sufficiently small $r$, the supports in
\begin{equation*}
 f_r(x)=\sum_i y_i\phi((x-x_i)/r)
\end{equation*}
are disjoint, so $f_r(x_i)=y_i$. If the test density is bounded by $M$ near those supports, change of variables gives
\begin{equation*}
 \|G_{f_r}\|_{\mathrm F}\leq\tr G_{f_r}
 =\E\|\nabla f_r(x)\|^2
 \leq M r^{D-2}\sum_i y_i^2\int\|\nabla\phi(u)\|^2\,du.
\end{equation*}
For $D>2$, this tends to zero although training interpolation is exact. The response's support has probability $O(r^D)$; for a bounded target, its test risk approaches the zero-predictor risk. This example concerns an unrestricted smooth function class and is not asserted to be the literal solution of our two-dimensional bottleneck model. It explains why a population-averaged sensitivity norm alone cannot identify memorization versus broadly useful feature learning.

\section{Reconstruction identity and its conditions}\label{app:tied}
For $f(x)=D_x(Hx+b)$ with $H=W^\top W$, select derivative zero at ReLU boundaries, matching automatic differentiation. The resulting selected derivative is $D_xH$ and $G=H(\E D_x)H$. A selection at a nondifferentiable point is not a classical derivative: for $f(x)=\max(x,0)$ and $X=0$ almost surely, this convention gives $G=0$ whereas $\E f(\epsilon Z)^2=\epsilon^2/2$. Thus Taylor-response statements do not automatically apply when the evaluation distribution charges ReLU boundaries. The following reconstruction identity remains valid, including on those boundaries. Define the test-set input energy $v=\E\|x\|^2$ and, for $\tau>0$, the local coefficients
\begin{equation}\label{eq:terms}
 q_{\rm gate}=\frac{\E\|D_xHx\|^2}{\tau},\quad
 B=\E[2\langle D_xHx,D_xb\rangle+\|D_xb\|^2],\quad
 C_x=\E\langle x,f(x)\rangle.
\end{equation}
The symbol $q_{\rm gate}$ is a scalar gate-dependent ratio, distinct from the token coordinates $q$ used in the language model. Expanding $\|x-f(x)\|^2$ gives
\begin{align*}
 L={}&\E\|x\|^2+\E\|D_xHx\|^2
 +\E[2\langle D_xHx,D_xb\rangle+\|D_xb\|^2]
 -2\E\langle x,f(x)\rangle.
\end{align*}
For $\tau>0$, substitute $\E\|D_xHx\|^2=q_{\rm gate}\tau$ and $\tau=\sqrt{r_{\rm eff}}F$ to obtain
\begin{equation}\label{eq:risk_decomp}
 L=v+q_{\rm gate}\sqrt{r_{\rm eff}}F+B-2C_x.
\end{equation} This is an algebraic identity, requiring finite moments but not gate--input independence. More explicitly, if $h_j=He_j$ and $M_j=\E[(D_x)_{jj}xx^\top]$, then $\E\|D_xHx\|^2=\sum_jh_j^\top M_jh_j$. Replacing $M_j$ by $\Pr((D_x)_{jj}=1)\E xx^\top$ would impose an additional independence assumption. Row normalization does not justify that replacement.

Across trained models indexed by $t$, write $a_t=q_{{\rm gate},t}\sqrt{r_{{\rm eff},t}}$ and $h_t=B_t-2C_{x,t}$. If $a_t$ and $h_t$ are approximately stable, $L_t=v+aF_t+h+e_t$ with small remainder $e_t=(a_t-a)F_t+(h_t-h)$. For centered vectors with $F_c\ne0$, if $\|e_c\|_2\leq\eta a\|F_c\|_2$ with $a>0$ and $0\leq\eta<1$, then
\begin{equation*}
 \operatorname{Corr}(L,F)\geq\sqrt{1-\eta^2}.
\end{equation*}
To see this, $L_c=aF_c+e_c$ has positive inner product with $aF_c$. The distance from $aF_c$ to the line through $L_c$ is at most $\|e_c\|_2$, bounding the sine of the acute angle by $\eta$. A large observed correlation alone does not establish stability of the mechanistic terms; this is a sufficient condition to measure.

\section{Perturbational interpretation and cross-entropy}\label{app:bridgeproof}
\subsection{Why the full norm describes directional amplification}
Let $G=U\operatorname{diag}(\lambda_i)U^\top$ and $z\sim\mathcal N(0,I_D)$. Orthogonal invariance gives $A(z)=\sum_i\lambda_i\xi_i^2$ with independent standard Gaussian $\xi_i$. Since $\E\xi_i^2=1$ and $\operatorname{Var}(\xi_i^2)=2$, $\E A=\sum_i\lambda_i=\tau$ and $\operatorname{Var}A=2\sum_i\lambda_i^2=2F^2$. With covariance $I_D/V$, these become $\tau/V$ and $2F^2/V^2$. The direction must be shared across examples before averaging: replacing $A(z)$ by independent directions per example estimates a different variance.

For a finite empirical probe set, the same identities apply exactly to $\widehat G=M^{-1}\sum_iJ_i^\top J_i$. A sample variance over finitely many Gaussian directions is a noisy estimator of $2\|\widehat G\|_{\mathrm F}^2/V^2$, distinct from the archived Rademacher matrix-vector estimate. Their agreement is a numerical check on the observable, not an independent generalization test.

The averaging order retains cross-example operator overlap. For independent $x,x'$ and finite required moments,
\begin{equation*}
 F^2=\E_{x,x'}\tr[(J_x^\top J_x)(J_{x'}^\top J_{x'})]
 =\E_{x,x'}\|J_xJ_{x'}^\top\|_{\mathrm F}^2.
\end{equation*}
In contrast, $\E_x\|J_x^\top J_x\|_{\mathrm F}^2$ averages single-example squares and generally differs. This functional overlap does not identify semantic feature overlap or establish predictive superiority over $\tau$.

Gaussian probes vary both orientation and length. For a uniform unit vector $u\in\mathbb S^{D-1}$, the sphere's second and fourth moments instead give
\begin{equation*}
 \E u^\top Gu=\tau/D,\qquad
 \operatorname{Var}(u^\top Gu)=\frac{2}{D(D+2)}\left(F^2-\frac{\tau^2}{D}\right).
\end{equation*}
For the mean squared nonlinear response, a sufficient averaging condition is
$\|f(x+\epsilon z)-f(x)-\epsilon J_f(x)z\|_{L^2(x,z)}=o(\epsilon)$ and $\|J_fz\|_{L^2}<\infty$; Cauchy--Schwarz then controls the cross term. These assumptions are stronger than specifying a derivative selection at boundary atoms.

\subsection{Probability changes and loss changes}
This supplementary calculation concerns local input perturbations, rather than the clean-loss association tested by the shape scan. With $\pi=\operatorname{softmax}(\ell)$ and $H_\pi=\operatorname{diag}(\pi)-\pi\pi^\top$, define
\begin{equation}\label{eq:fisher}
 K=\E[J_\ell^\top H_\pi J_\ell],\qquad
 0\preceq K\preceq\tfrac12s_\ell^2G.
\end{equation}
For independent $z\sim\mathcal N(0,I_{CV}/V)$ and $\delta=\epsilon z$, twice differentiable logits with composite Taylor remainders $o(\epsilon^2)$ in $L^1$ give
\begin{align}\label{eq:lm_bridge}
 \E_{q,z}\operatorname{KL}(\pi(q)\|\pi(q+\delta))
 &=\frac{\epsilon^2}{2V}\tr K+o(\epsilon^2),\nonumber\\
 \E_{q,y,z}[\ell_{\rm CE}(q+\delta,y)-\ell_{\rm CE}(q,y)]
 &=\frac{\epsilon^2}{2V}\big[\underbrace{\omega s_\ell^2\sqrt{r_{\rm eff}}F}_{\tr K}+\Gamma\big]+o(\epsilon^2).
\end{align}
Here $\Gamma=\E\sum_k(\pi_k-\mathbf1\{y=k\})\tr(\nabla_q^2\ell_k)$ and, for $\tau>0$, $\omega=\tr K/(s_\ell^2\tau)$. These auxiliary quantities describe the probability weighting and nonlinear curvature needed to interpret a CE perturbation. The derivation follows.
Write $J_\ell=D\ell$, $\pi=\operatorname{softmax}(\ell)$, and $H_\pi=\operatorname{diag}(\pi)-\pi\pi^\top$. The Hessian of $\log\sum_k\exp\ell_k$ is $H_\pi$, so a local logit displacement $\Delta\ell$ satisfies
\begin{equation*}
 \operatorname{KL}(\pi(\ell)\|\pi(\ell+\Delta\ell))
 =\tfrac12\Delta\ell^\top H_\pi\Delta\ell+o(\|\Delta\ell\|^2).
\end{equation*}
Substitution of $\Delta\ell=J_\ell\delta+o(\|\delta\|)$ and $\E\delta\delta^\top=\epsilon^2I_D/V$ gives the first line of Equation~\eqref{eq:lm_bridge}. Assume twice differentiable logits and $o(\epsilon^2)$ remainders in $L^1(q,y,z)$ for the composite KL and CE expansions. These conditions explicitly justify averaging; finite moments of locally dominating derivatives are one sufficient route.

For a one-hot label $e_y$, the exact Hessian decomposition is
\begin{equation*}
 \nabla_q^2\ell_{\rm CE}(q,y)
 =J_\ell^\top H_\pi J_\ell+\sum_k(\pi_k-(e_y)_k)\nabla_q^2\ell_k.
\end{equation*}
A symmetric perturbation cancels first-order terms. Averaging the second-order quadratic form proves the CE expansion, with $\Gamma$ the trace of the second term. It vanishes for locally affine logits, and in conditional expectation when labels follow the model distribution; neither condition is assumed in the audit. $\Gamma$ can have either sign and is a second-order contribution, not a higher-order Taylor remainder.

To bound the first term, $H_\pi\mathbf1=0$ and $H_\pi$ is positive semidefinite. Its row diagonal and sum of absolute off-diagonal entries are both $\pi_i(1-\pi_i)$, so its spectral norm is at most $\max_i2\pi_i(1-\pi_i)\leq1/2$. Therefore
\begin{equation*}
 0\preceq K\preceq\tfrac12\E[J_\ell^\top C_VJ_\ell]
 =\tfrac12s_\ell^2G,\qquad
 \|K\|_{\mathrm F}\leq\tfrac12s_\ell^2F.
\end{equation*}
The last inequality follows because for $0\preceq A\preceq B$, $\tr(B^2-A^2)=\tr[(B-A)(B+A)]\geq0$. With $s_\ell>0$ and $\tau>0$, $\omega=\tr K/(s_\ell^2\tau)\in[0,1/2]$. If $\tau=0$, the Fisher contribution is zero without defining $\omega$. Combining these identities gives Equation~\eqref{eq:lm_bridge}. Output scale, effective rank, probability weighting, and nonlinear logit curvature can each affect a loss ranking.

\section{A conditional connection to clean cross-entropy}\label{app:cleanbridge}
This is a sufficient mathematical model of joint growth, not a decomposition identified in the Transformer checkpoints.
\begin{proposition}[Conditional common growth]\label{thm:clean}
Choose smooth centered logits $\ell_a=\ell_*+ah$, $a\geq0$, on the evaluation neighborhoods, with $\operatorname{softmax}(\ell_*)=p_*>0$ on the support, finite required moments, and differentiation under expectation justified. If $\E[J_*^\top J_h+J_h^\top J_*]=0$, then raw AGOP satisfies
\begin{equation}\label{eq:cleangeometry}
 \|\widetilde G_a\|_{\mathrm F}^2
 =\|G_*\|_{\mathrm F}^2+2a^2\tr(G_*G_h)+a^4\|G_h\|_{\mathrm F}^2.
\end{equation}
Clean CE and raw norm are nondecreasing in $a$. Both strictly increase for $a>0$ if $G_h\ne0$ and $h$ has nonconstant class components on a positive-probability set.
\end{proposition}
Assume positive true class probabilities on the evaluation support. Choose smooth centered-logit extensions $\ell_*,h$ on its probe neighborhoods, with $\operatorname{softmax}(\ell_*)=p_*$ on the support, and consider $\ell_a=\ell_*+ah$, $a\geq0$. For discrete token inputs, Bayes probabilities specify values only on the discrete support; they do not determine this extension or its derivatives. Every AGOP statement here is relative to the selected extension. Assume finite gradient second moments, $\E\|h\|^3<\infty$, and sufficient domination to differentiate the expectations below. All operators initially use raw centered logits. The identity
\begin{equation}\label{eq:clean_ce}
 L_a-H(Y\mid Q)=\E_Q\operatorname{KL}(p_*(\cdot\mid Q)\|p_a(\cdot\mid Q))
\end{equation}
follows by adding and subtracting $-\log p_*(y\mid q)$. Write $A(\ell)=\log\sum_k e^{\ell_k}$. Exactly,
\begin{align*}
 L(a)-H&=\E[A(\ell_*+ah)-A(\ell_*)-a p_*^\top h],\\
 L'(0)&=0,\qquad L''(a)=\E[h^\top H_{p_a}h]\geq0.
\end{align*}
Thus $L(a)$ is nondecreasing for all $a\geq0$, without a small-error approximation. It is strictly increasing for $a>0$ when $h$ has nonconstant class components on a set of positive probability. Positive softmax probabilities make $H_{p_a}$ positive definite on centered class directions. A Taylor expansion additionally gives, with $H_*=\operatorname{diag}(p_*)-p_*p_*^\top$,
\begin{equation}\label{eq:clean_expand}
 L_a-H(Y\mid Q)=\tfrac12 a^2 C_h+O(a^3\E\|h\|^3),
 \qquad C_h=\E[h(q)^\top H_*(q)h(q)]\geq0.
\end{equation}
The expansion describes approximate proportionality at small logit error, not the exact monotonicity just proved.

Let $J_*=D\ell_*$ and $J_h=Dh$. Then
\begin{equation*}
 \widetilde G_a=G_*+a\mathcal C+a^2G_h,\qquad
 \mathcal C=\E[J_*^\top J_h+J_h^\top J_*].
\end{equation*}
If $\mathcal C=0$, squaring the operator gives Equation~\eqref{eq:cleangeometry}. Since $\tr(G_*G_h)\geq0$, its norm is strictly increasing for $a>0$ when $G_h\ne0$. Together with the nonconstant-class condition, this proves Proposition~\ref{thm:clean}. Both nondegeneracy conditions are necessary: a constant-in-input class bias can change CE with $G_h=0$, whereas a function vanishing on discrete evaluation points can have nonzero derivatives and no clean CE effect.

The reverse triangle inequality also permits a nonzero cross operator:
\begin{equation*}
 \left|\|\widetilde G_a\|_{\mathrm F}-a^2\|G_h\|_{\mathrm F}\right|
 \leq\|G_*\|_{\mathrm F}+a\|\mathcal C\|_{\mathrm F}.
\end{equation*}
If the CE Taylor residual is bounded by $K a^3\E\|h\|^3$, then for $G_h\ne0$,
\begin{align}\label{eq:clean_bound}
 \left|L_a-H-\frac{C_h}{2\|G_h\|_{\mathrm F}}s_a^2F_a\right|
 \leq K a^3\E\|h\|^3+
 \frac{C_h}{2\|G_h\|_{\mathrm F}}\big(\|G_*\|_{\mathrm F}+a\|\mathcal C\|_{\mathrm F}\big).
\end{align}
Here $s_a>0$ and $\|\widetilde G_a\|_{\mathrm F}=s_a^2F_a$. This bound states explicitly the task-gradient, cross-gradient, and Taylor terms that must be small relative to the leading error contribution. For a fixed nonzero $G_*$, taking $a\to0$ does not make $a^2G_h$ dominant. A possible two-parameter regime uses $h_k(q)=v\sin(kq_1)$ on a continuous uniform coordinate, $k\to\infty$, and $a=k^{-1/2}$. Its logit amplitude vanishes while $a^2\|G_{h_k}\|_{\mathrm F}$ grows as $k$; if $G_*$ is bounded and the cross term is controlled, the relative errors above vanish. This illustrates compatibility of the conditions, not a fitted model of language errors.

Normalization is substantive. If $\ell_*=0$ and $\ell_a=ah$ with $a>0$, then $s_a^2=a^2s_h^2$ and $F_a=\|G_h\|_{\mathrm F}/s_h^2$ is constant even when raw CE increases. Hence common raw-norm growth does not imply normalized-norm growth without scale control. We report both raw and normalized measurements.

These are sufficient mechanisms, not an inferred decomposition of the corpus's Bayes logits. Shared error directions, selected smooth extensions and controlled cross operators have not been identified in the Transformer scan. The local perturbation experiment tests an observable response consequence; it does not validate these latent clean-loss assumptions.

\section{A conditional width--depth allocation model}\label{app:allocation}
Following the representation--processing distinction of \citet{liu2025}, suppose
\begin{equation*}
 L(P,T,d,w)=E(T)+A_w w^{-\beta}+A_d d^{-\alpha},\qquad
 A_w,A_d,\alpha,\beta>0.
\end{equation*}
With fixed coefficients and $P=\kappa dw^2$, write $u=\log d$. At fixed $P,T$,
\begin{equation*}
 L-E(T)=A_w(P/\kappa)^{-\beta/2}e^{\beta u/2}+A_de^{-\alpha u}.
\end{equation*}
This is strictly convex in $u$ and diverges at both ends, so its unique continuous minimum satisfies
\begin{equation*}
 \tfrac\beta2 A_w w_*^{-\beta}=\alpha A_d d_*^{-\alpha},\qquad
 d_*\propto P^{\beta/(2\alpha+\beta)},\quad
 w_*\propto P^{\alpha/(2\alpha+\beta)}.
\end{equation*}
The optimum balances marginal improvements, rather than requiring equal loss terms. For inverse-width and inverse-depth terms ($\alpha=\beta=1$), both dimensions scale as $P^{1/3}$ and $w_*/d_*=A_w/(2A_d)$, consistent with the coefficient-based analysis of \citet{liu2026coefficients}. Our empirical fit permits a changing center under its actual training recipe. This distinction does not falsify the conditional calculation: discretization, FFN ratios, overhead, data exposure, and optimization can all change the observed optimum. Neither fit separately measures a latent processing-error term.

\section{Experimental protocols and provenance}\label{app:protocol}
Historical results are recomputed from saved measurements, with per-checkpoint norms calculated before averaging seeds. The revision retains the original measurements, source snapshots, and checksums. The analysis scripts generate tables and figures from these files. No old off-diagonal plot is relabeled as a full-norm result.

\subsection{Sparse reconstruction}
The complete sweep has 22 sample sizes and five initializations per size. AdamW uses learning rate 0.005, weight decay 0.01, batch size $\min(n,2048)$, 25\% warmup, and cosine decay over 3,000 steps. Dataset seeds depend on sample size; the five initializations share its generated train/test datasets. Initialization SEM therefore does not quantify data-sampling uncertainty. The empirical input AGOP is $HP_DH$, evaluated using test activation frequencies. Saved fields contain off-diagonal squared energy $E_{\rm off}$ and its fraction $\rho_{\rm off}=E_{\rm off}/F^2$. Every fraction is positive, permitting recovery of $F=\sqrt{E_{\rm off}/\rho_{\rm off}}$ separately for each run. This inherits the precision of the archived fields. The zero-predictor population risk is $1-0.99^{1000}\simeq0.99996$ because normalized nonzero inputs have unit energy. Table~\ref{tab:ddall} reports the full range used in Figure~\ref{fig:dd}.
\begin{table}[htbp]\centering\small
\begin{tabular}{rrrr}\toprule
$n$ & Train loss & Test loss & $F$ \\ \midrule
3 & 0.000000 & 1.102627 & 2.708933 \\
5 & 0.040151 & 0.998945 & 0.528551 \\
8 & 0.139483 & 1.007707 & 0.254106 \\
10 & 0.288048 & 1.007314 & 0.551626 \\
15 & 0.475842 & 1.003641 & 0.387439 \\
30 & 0.689330 & 1.010302 & 5.708989 \\
50 & 0.801488 & 1.009268 & 6.636263 \\
100 & 0.880191 & 1.032078 & 41.972037 \\
200 & 0.931517 & 1.248293 & 379.912159 \\
500 & 0.960633 & 1.534006 & 670.396083 \\
1000 & 0.975411 & 1.058263 & 54.598458 \\
1395 & 0.980733 & 1.000026 & 8.420884 \\
1946 & 0.983593 & 0.994491 & 2.872093 \\
2714 & 0.985116 & 0.993201 & 2.192568 \\
3786 & 0.987104 & 0.993174 & 2.149345 \\
5282 & 0.988158 & 0.992724 & 2.087510 \\
7368 & 0.989383 & 0.993290 & 1.965353 \\
10278 & 0.990260 & 0.992984 & 1.994001 \\
14337 & 0.991034 & 0.993052 & 2.062209 \\
20000 & 0.991313 & 0.993113 & 1.909867 \\
30000 & 0.992067 & 0.993586 & 2.098628 \\
40000 & 0.992337 & 0.993247 & 2.184659 \\
\bottomrule\end{tabular}

\caption{Complete sparse-reconstruction sweep, means of five initialization seeds at each training size.}\label{tab:ddall}
\end{table}

\subsection{Language-model corpus and exposure}
The corpus is derived from a FineWeb \texttt{sample-10BT} shard encoded as UTF-8 bytes. The archived adaptive runs use a 3.08B-byte training split and approximately 67M-byte validation and test splits. Their nominal exposure target is $60P$; convergence audits can extend exposure and restore a best-validation state. Table~\ref{tab:allshapes} records processed-byte exposure, including repeats and optimization work before any restoration. The candidate grid begins at depths $\{1,2,3,4,5,6,8,10,12,16,20,24\}$, with larger depths and additional seeds supplied by adaptive extensions.

The base optimizer is AdamW with learning rate $3\times10^{-4}$, weight decay 0.01, batch size 64, context length 256, gradient clipping at norm one, and 300 warmup steps followed by cosine decay. Saved per-run configurations record the actual schedules. Corpus and checkpoint metadata for the 15M rows are given with Table~\ref{tab:allshapes}. Every budget uses seed-mean selection-validation CE to choose its frontier point and contributes equal total weight to the allocation fit. The fitting analysis uses the available configuration means; archived training-time selection decisions remain in the underlying records.

\section{AGOP estimation in language models}\label{app:estimator}
The standard output-by-input Jacobian convention throughout this paper gives $G=\E[J_f^\top J_f]$. For $M$ fixed examples, $\widehat G=M^{-1}\sum_iJ_i^\top J_i$. Shared independent Rademacher probes $z_r\in\{-1,1\}^{CV}$ yield
\begin{equation*}
 \widehat Q=R^{-1}\sum_{r=1}^R\|\widehat Gz_r\|^2,\qquad
 \E_z[\widehat Q\mid q_1,\ldots,q_M]=\|\widehat G\|_{\mathrm F}^2.
\end{equation*}
Jacobian-vector and vector-Jacobian products evaluate this operator without forming its $65{,}536\times65{,}536$ matrix. Matrix-vector products are averaged across examples before squaring. The reported $F=\sqrt{\widehat Q}$ is not an exactly unbiased norm estimator. The main measurements use $M=R=96$; auxiliary output probes estimate the trace. Probe variation, example variation, and training-seed variation are distinct uncertainty sources.

Conditional probe unbiasedness differs from population unbiasedness. For a fixed function and output scale (or raw logits), iid examples with $A_i=J_i^\top J_i$ and finite second operator moments give
\begin{equation*}
 \E_{\rm data}\|\widehat G\|_{\mathrm F}^2=\|G\|_{\mathrm F}^2+\frac1M\E\|A-G\|_{\mathrm F}^2.
\end{equation*}
The cross-example estimator $U=[M(M-1)]^{-1}\sum_{i\ne j}\tr(A_iA_j)$ is unbiased for population $F^2$ under those assumptions. One isotropic probe gives the unbiased estimator
\begin{equation*}
 U_z=\frac{\|\sum_i A_i z\|^2-\sum_i\|A_i z\|^2}{M(M-1)}.
\end{equation*}
Finite-probe estimates can be negative, and square roots restore bias. Estimating the RMS on the same examples introduces additional scale dependence, so the raw-operator identity does not imply unbiasedness after that normalization. Finite-corpus windows need not be iid population draws. Appendix~\ref{app:newaudit} separately increases $M$ and $R$ and reports $U_z$; this does not establish population convergence.

The ambient continuous extension of one-hot inputs is not intrinsic to discrete predictions. Replacing $E$ by $E+\mathbf1c^\top$ and $p_t$ by $p_t-c$ leaves $q_tE+p_t$ unchanged whenever $q_t\mathbf1=1$, but changes its derivative in simplex-normal directions. The tangent operator
\begin{equation*}
 P_T=I_C\otimes(I_V-\mathbf1\mathbf1^\top/V),\qquad G_T=P_TGP_T
\end{equation*}
removes this particular normal-direction freedom. It still depends on a continuous extension and does not measure natural-language edit robustness. Centering removes common-logit shifts, RMS is fixed when differentiating, and the embedding matrix remains inside the function. Main-scan CE averages token positions; the local audit aligns last-position CE with AGOP. The archived estimator replaces nonfinite derivatives with zero without recording counts. New checks reject nonfinite values, but cannot retrospectively certify every historical scalar.

\section{Complete language-model tables}\label{app:lmtables}
All repeated initializations are averaged within identical $(P_0,d,w,m,h)$ configurations. Norms are averaged as norms before taking logs. The table contains all 152 configuration means across twelve budgets. Head dimension is four; head count and FFN ratios change with width to match parameters. Exposure $T$ counts repeated bytes.
\begingroup\scriptsize\setlength{\tabcolsep}{2.6pt}
\begin{longtable}{rrrrrrrrrrrr}
\caption{All 152 configuration means across twelve budgets (254 runs). $P_0$, active $P$ and processed byte exposure $T$ are in millions. $d,w,m,h,a$ denote depth, residual width, FFN width, head count and head dimension; $s$ counts seeds. Norms are averaged before taking logs.}\label{tab:allshapes}\\
\toprule $P_0$ & $P$ & $T$ & $d$ & $w$ & $m$ & $h$ & $a$ & $s$ & Val CE & Test CE & $\log F$ \\
\midrule\endfirsthead
\multicolumn{12}{l}{\textit{Table \thetable\ continued}}\\
\toprule $P_0$ & $P$ & $T$ & $d$ & $w$ & $m$ & $h$ & $a$ & $s$ & Val CE & Test CE & $\log F$ \\
\midrule\endhead\bottomrule\endfoot
0.3 & 0.300032 & 73.76 & 1 & 128 & 529 & 32 & 4 & 1 & 1.96654 & 1.99316 & 8.967 \\
0.3 & 0.300096 & 73.76 & 2 & 96 & 395 & 24 & 4 & 1 & 1.96018 & 1.98780 & 9.512 \\
0.3 & 0.300048 & 69.26 & 3 & 84 & 297 & 21 & 4 & 4 & 1.96063 & 1.98731 & 9.700 \\
0.3 & 0.300240 & 73.83 & 4 & 72 & 279 & 18 & 4 & 1 & 1.94698 & 1.97405 & 9.627 \\
0.3 & 0.300160 & 60.32 & 5 & 64 & 262 & 16 & 4 & 4 & 2.01452 & 2.04017 & 9.551 \\
0.3 & 0.300360 & 73.83 & 6 & 60 & 231 & 15 & 4 & 1 & 1.94650 & 1.97391 & 9.614 \\
0.3 & 0.299624 & 60.22 & 8 & 52 & 206 & 13 & 4 & 4 & 2.05792 & 2.08299 & 9.733 \\
0.3 & 0.300000 & 73.76 & 10 & 48 & 176 & 12 & 4 & 1 & 2.00285 & 2.02898 & 9.205 \\
0.3 & 0.299992 & 73.76 & 12 & 44 & 162 & 11 & 4 & 1 & 2.02794 & 2.05468 & 9.746 \\
0.3 & 0.299592 & 73.70 & 16 & 36 & 162 & 9 & 4 & 1 & 2.06724 & 2.09223 & 10.031 \\
0.3 & 0.298560 & 73.43 & 20 & 32 & 148 & 8 & 4 & 1 & 2.10576 & 2.13039 & 10.182 \\
0.3 & 0.299584 & 73.70 & 24 & 32 & 113 & 8 & 4 & 1 & 2.12691 & 2.15016 & 9.885 \\
0.6 & 0.600000 & 147.59 & 1 & 200 & 713 & 50 & 4 & 1 & 1.65269 & 1.68443 & 10.041 \\
0.6 & 0.599992 & 147.59 & 2 & 148 & 523 & 37 & 4 & 1 & 1.63423 & 1.66593 & 10.150 \\
0.6 & 0.600000 & 147.59 & 3 & 120 & 463 & 30 & 4 & 1 & 1.58254 & 1.61584 & 9.346 \\
0.6 & 0.600080 & 120.59 & 4 & 104 & 415 & 26 & 4 & 4 & 1.60388 & 1.63609 & 9.546 \\
0.6 & 0.600000 & 120.59 & 5 & 96 & 354 & 24 & 4 & 4 & 1.59589 & 1.62793 & 9.367 \\
0.6 & 0.599984 & 120.59 & 6 & 88 & 326 & 22 & 4 & 4 & 1.60124 & 1.63301 & 9.400 \\
0.6 & 0.600336 & 147.65 & 8 & 72 & 327 & 18 & 4 & 1 & 1.58674 & 1.62011 & 9.747 \\
0.6 & 0.600440 & 147.65 & 10 & 68 & 265 & 17 & 4 & 1 & 1.59457 & 1.62572 & 9.688 \\
0.6 & 0.600600 & 147.72 & 12 & 60 & 263 & 15 & 4 & 1 & 1.63082 & 1.66246 & 9.697 \\
0.6 & 0.600432 & 147.65 & 16 & 56 & 197 & 14 & 4 & 1 & 1.63229 & 1.66480 & 9.294 \\
0.6 & 0.599520 & 147.46 & 20 & 48 & 195 & 12 & 4 & 1 & 1.65797 & 1.68920 & 9.751 \\
0.6 & 0.599896 & 147.52 & 24 & 44 & 178 & 11 & 4 & 1 & 1.68830 & 1.71974 & 9.431 \\
1 & 1.000032 & 96.01 & 1 & 264 & 979 & 66 & 4 & 1 & 1.67256 & 1.70307 & 6.952 \\
1 & 0.999992 & 96.01 & 2 & 196 & 689 & 49 & 4 & 1 & 1.66243 & 1.69359 & 9.165 \\
1 & 0.999960 & 95.99 & 3 & 156 & 626 & 39 & 4 & 1 & 1.58014 & 1.61311 & 8.950 \\
1 & 1.000144 & 96.01 & 4 & 136 & 549 & 34 & 4 & 4 & 1.56639 & 1.59876 & 9.002 \\
1 & 0.999680 & 95.94 & 5 & 128 & 446 & 32 & 4 & 4 & 1.55660 & 1.58929 & 8.950 \\
1 & 1.000160 & 96.01 & 6 & 112 & 454 & 28 & 4 & 4 & 1.56835 & 1.60026 & 8.957 \\
1 & 1.000128 & 96.01 & 8 & 96 & 409 & 24 & 4 & 1 & 1.57913 & 1.61134 & 9.431 \\
1 & 1.000040 & 96.01 & 10 & 92 & 319 & 23 & 4 & 1 & 1.57424 & 1.60620 & 9.120 \\
1 & 1.000104 & 96.01 & 12 & 84 & 294 & 21 & 4 & 1 & 1.59830 & 1.62984 & 8.907 \\
1 & 1.000080 & 96.01 & 16 & 72 & 264 & 18 & 4 & 1 & 1.62549 & 1.65825 & 9.094 \\
1 & 0.999040 & 95.90 & 20 & 64 & 241 & 16 & 4 & 1 & 1.66800 & 1.69836 & 8.808 \\
1 & 1.000048 & 96.01 & 24 & 56 & 242 & 14 & 4 & 1 & 1.71391 & 1.74339 & 9.262 \\
1.6 & 1.599984 & 153.60 & 1 & 328 & 1396 & 82 & 4 & 1 & 1.53981 & 1.57066 & 8.369 \\
1.6 & 1.600152 & 153.60 & 2 & 244 & 957 & 61 & 4 & 1 & 1.53215 & 1.56308 & 9.124 \\
1.6 & 1.600000 & 153.60 & 3 & 200 & 803 & 50 & 4 & 1 & 1.44339 & 1.47459 & 8.695 \\
1.6 & 1.599840 & 153.57 & 4 & 176 & 686 & 44 & 4 & 1 & 1.43490 & 1.46628 & 8.760 \\
1.6 & 1.600000 & 153.60 & 5 & 160 & 601 & 40 & 4 & 4 & 1.42498 & 1.45567 & 8.838 \\
1.6 & 1.599640 & 153.57 & 6 & 140 & 606 & 35 & 4 & 4 & 1.43070 & 1.46141 & 8.924 \\
1.6 & 1.599744 & 153.57 & 8 & 128 & 475 & 32 & 4 & 4 & 1.43112 & 1.46154 & 8.754 \\
1.6 & 1.599640 & 153.57 & 10 & 116 & 417 & 29 & 4 & 1 & 1.43729 & 1.46788 & 8.818 \\
1.6 & 1.600144 & 153.60 & 12 & 104 & 399 & 26 & 4 & 1 & 1.44721 & 1.47704 & 8.656 \\
1.6 & 1.599664 & 153.57 & 16 & 88 & 366 & 22 & 4 & 1 & 1.47494 & 1.50482 & 9.025 \\
1.6 & 1.600200 & 153.62 & 20 & 84 & 287 & 21 & 4 & 1 & 1.47552 & 1.50648 & 8.660 \\
1.6 & 1.600272 & 153.62 & 24 & 72 & 301 & 18 & 4 & 1 & 1.50214 & 1.53283 & 8.623 \\
2.3 & 2.300000 & 220.79 & 1 & 400 & 1688 & 100 & 4 & 1 & 1.47184 & 1.50199 & 8.363 \\
2.3 & 2.299968 & 220.79 & 2 & 288 & 1226 & 72 & 4 & 1 & 1.45880 & 1.48905 & 9.010 \\
2.3 & 2.300048 & 220.82 & 3 & 232 & 1058 & 58 & 4 & 1 & 1.37656 & 1.40724 & 8.586 \\
2.3 & 2.300064 & 220.82 & 4 & 208 & 868 & 52 & 4 & 1 & 1.36552 & 1.39559 & 8.886 \\
2.3 & 2.300000 & 220.79 & 5 & 184 & 803 & 46 & 4 & 4 & 1.35640 & 1.38602 & 8.635 \\
2.3 & 2.300320 & 220.84 & 6 & 176 & 671 & 44 & 4 & 4 & 1.35503 & 1.38478 & 8.590 \\
2.3 & 2.299680 & 220.77 & 8 & 144 & 660 & 36 & 4 & 1 & 1.37191 & 1.40169 & 8.684 \\
2.3 & 2.300200 & 220.82 & 10 & 140 & 501 & 35 & 4 & 4 & 1.36424 & 1.39378 & 8.348 \\
2.3 & 2.300696 & 220.87 & 12 & 124 & 491 & 31 & 4 & 1 & 1.37142 & 1.40049 & 8.332 \\
2.3 & 2.299856 & 220.79 & 16 & 104 & 457 & 26 & 4 & 1 & 1.37513 & 1.40447 & 8.314 \\
2.3 & 2.300920 & 220.89 & 20 & 92 & 420 & 23 & 4 & 1 & 1.39381 & 1.42403 & 8.227 \\
2.3 & 2.300920 & 220.89 & 24 & 92 & 319 & 23 & 4 & 1 & 1.39168 & 1.42149 & 8.377 \\
3 & 3.000000 & 288.00 & 1 & 480 & 1778 & 120 & 4 & 1 & 1.43629 & 1.46545 & 8.813 \\
3 & 2.999888 & 287.98 & 2 & 328 & 1436 & 82 & 4 & 1 & 1.41598 & 1.44543 & 8.649 \\
3 & 3.000008 & 288.00 & 3 & 292 & 998 & 73 & 4 & 1 & 1.33101 & 1.36077 & 8.674 \\
3 & 2.999760 & 287.98 & 4 & 232 & 1054 & 58 & 4 & 1 & 1.32706 & 1.35693 & 8.663 \\
3 & 2.999800 & 287.98 & 5 & 212 & 912 & 53 & 4 & 4 & 1.31358 & 1.34175 & 8.382 \\
3 & 3.000192 & 288.03 & 6 & 192 & 852 & 48 & 4 & 4 & 1.31319 & 1.34125 & 8.402 \\
3 & 3.000024 & 288.00 & 8 & 172 & 696 & 43 & 4 & 1 & 1.31705 & 1.34584 & 8.187 \\
3 & 3.000000 & 288.00 & 10 & 160 & 577 & 40 & 4 & 4 & 1.31298 & 1.34120 & 7.994 \\
3 & 3.000096 & 288.00 & 12 & 144 & 546 & 36 & 4 & 1 & 1.32071 & 1.34961 & 7.776 \\
3 & 3.000056 & 288.00 & 16 & 124 & 482 & 31 & 4 & 1 & 1.32655 & 1.35524 & 7.821 \\
3 & 2.999160 & 287.93 & 20 & 108 & 457 & 27 & 4 & 1 & 1.33544 & 1.36389 & 7.719 \\
3 & 3.000000 & 288.00 & 24 & 96 & 441 & 24 & 4 & 1 & 1.34896 & 1.37816 & 7.667 \\
5 & 5.000016 & 479.99 & 1 & 644 & 2207 & 161 & 4 & 1 & 1.38631 & 1.41409 & 8.050 \\
5 & 5.000024 & 479.99 & 2 & 452 & 1667 & 113 & 4 & 1 & 1.34390 & 1.37126 & 8.002 \\
5 & 5.000048 & 479.99 & 3 & 376 & 1334 & 94 & 4 & 1 & 1.26672 & 1.29399 & 8.056 \\
5 & 5.000072 & 479.99 & 4 & 308 & 1315 & 77 & 4 & 1 & 1.25656 & 1.28355 & 7.857 \\
5 & 4.999680 & 479.97 & 5 & 288 & 1081 & 72 & 4 & 1 & 1.23806 & 1.26391 & 7.795 \\
5 & 5.000176 & 480.03 & 6 & 248 & 1118 & 62 & 4 & 1 & 1.23885 & 1.26490 & 7.892 \\
5 & 4.999896 & 479.99 & 8 & 236 & 802 & 59 & 4 & 4 & 1.23407 & 1.25987 & 7.585 \\
5 & 5.000040 & 479.99 & 10 & 204 & 777 & 51 & 4 & 4 & 1.23534 & 1.26123 & 7.285 \\
5 & 5.000064 & 479.99 & 12 & 192 & 667 & 48 & 4 & 4 & 1.23540 & 1.26124 & 7.173 \\
5 & 5.000496 & 480.05 & 16 & 152 & 698 & 38 & 4 & 1 & 1.24105 & 1.26679 & 7.147 \\
5 & 5.000720 & 480.08 & 20 & 136 & 626 & 34 & 4 & 1 & 1.24683 & 1.27349 & 7.081 \\
5 & 4.999368 & 479.94 & 24 & 132 & 507 & 33 & 4 & 1 & 1.24505 & 1.27149 & 6.733 \\
10 & 10.000048 & 960.02 & 1 & 904 & 3336 & 226 & 4 & 1 & 1.33344 & 1.35943 & 8.497 \\
10 & 10.000144 & 960.02 & 2 & 616 & 2632 & 154 & 4 & 1 & 1.23941 & 1.26300 & 7.310 \\
10 & 10.000000 & 960.02 & 3 & 500 & 2203 & 125 & 4 & 1 & 1.18761 & 1.21233 & 7.335 \\
10 & 10.000080 & 960.02 & 4 & 456 & 1731 & 114 & 4 & 1 & 1.16758 & 1.19113 & 7.110 \\
10 & 10.000000 & 960.02 & 5 & 400 & 1621 & 100 & 4 & 1 & 1.15643 & 1.17912 & 7.027 \\
10 & 10.000104 & 960.02 & 6 & 372 & 1430 & 93 & 4 & 1 & 1.15074 & 1.17320 & 6.887 \\
10 & 10.000000 & 960.02 & 8 & 320 & 1263 & 80 & 4 & 1 & 1.14596 & 1.16874 & 6.799 \\
10 & 9.999640 & 959.95 & 10 & 284 & 1152 & 71 & 4 & 1 & 1.14026 & 1.16245 & 6.406 \\
10 & 9.998736 & 959.87 & 12 & 264 & 1016 & 66 & 4 & 4 & 1.14160 & 1.16424 & 5.851 \\
10 & 9.999808 & 959.97 & 16 & 224 & 921 & 56 & 4 & 4 & 1.14129 & 1.16362 & 5.596 \\
10 & 10.000040 & 960.02 & 20 & 212 & 734 & 53 & 4 & 4 & 1.13799 & 1.16046 & 5.589 \\
10 & 9.999744 & 959.97 & 24 & 192 & 683 & 48 & 4 & 1 & 1.14181 & 1.16392 & 5.609 \\
15 & 15.000120 & 1440.01 & 4 & 540 & 2294 & 135 & 4 & 2 & 1.12091 & 1.11595 & 7.288 \\
15 & 15.000120 & 1440.01 & 8 & 380 & 1657 & 95 & 4 & 2 & 1.09139 & 1.08614 & 6.640 \\
15 & 14.999712 & 1440.01 & 12 & 336 & 1154 & 84 & 4 & 2 & 1.08643 & 1.08129 & 5.167 \\
15 & 14.999960 & 1440.01 & 16 & 268 & 1187 & 67 & 4 & 2 & 1.08607 & 1.08121 & 4.989 \\
15 & 14.999488 & 1440.01 & 24 & 224 & 929 & 56 & 4 & 2 & 1.08439 & 1.07963 & 4.809 \\
15 & 15.001752 & 1440.01 & 32 & 204 & 727 & 51 & 4 & 2 & 1.08443 & 1.07951 & 4.880 \\
15 & 14.999376 & 1440.01 & 48 & 168 & 584 & 42 & 4 & 2 & 1.08669 & 1.08216 & 5.038 \\
20 & 20.000024 & 1920.01 & 1 & 1276 & 4898 & 319 & 4 & 1 & 1.29128 & 1.31695 & 8.081 \\
20 & 19.999944 & 1919.99 & 2 & 916 & 3432 & 229 & 4 & 1 & 1.14548 & 1.16574 & 7.653 \\
20 & 20.000024 & 1920.01 & 3 & 748 & 2830 & 187 & 4 & 1 & 1.12019 & 1.14092 & 6.220 \\
20 & 20.000000 & 1920.01 & 4 & 640 & 2528 & 160 & 4 & 1 & 1.09292 & 1.11383 & 6.917 \\
20 & 19.999440 & 1919.94 & 5 & 564 & 2339 & 141 & 4 & 1 & 1.08475 & 1.10406 & 6.411 \\
20 & 20.000240 & 1920.04 & 6 & 520 & 2099 & 130 & 4 & 1 & 1.07478 & 1.09471 & 6.351 \\
20 & 19.999616 & 1919.94 & 8 & 448 & 1844 & 112 & 4 & 1 & 1.06963 & 1.08844 & 5.542 \\
20 & 20.000080 & 1920.01 & 10 & 424 & 1470 & 106 & 4 & 1 & 1.06367 & 1.08298 & 4.971 \\
20 & 19.999488 & 1919.94 & 12 & 384 & 1368 & 96 & 4 & 1 & 1.06456 & 1.08413 & 3.770 \\
20 & 20.000520 & 1920.06 & 16 & 324 & 1255 & 81 & 4 & 1 & 1.06039 & 1.07934 & 4.959 \\
20 & 20.000720 & 1920.06 & 20 & 296 & 1076 & 74 & 4 & 4 & 1.05665 & 1.07554 & 4.299 \\
20 & 20.002336 & 1920.20 & 24 & 272 & 970 & 68 & 4 & 4 & 1.05663 & 1.07592 & 4.755 \\
20 & 20.000720 & 1920.06 & 32 & 232 & 869 & 58 & 4 & 1 & 1.05801 & 1.07673 & 4.913 \\
20 & 20.001800 & 1920.16 & 40 & 196 & 872 & 49 & 4 & 4 & 1.05873 & 1.07798 & 4.509 \\
20 & 19.999104 & 1919.89 & 48 & 192 & 691 & 48 & 4 & 1 & 1.05861 & 1.07807 & 4.238 \\
20 & 19.999056 & 1919.89 & 64 & 168 & 586 & 42 & 4 & 1 & 1.06325 & 1.08292 & 4.958 \\
30 & 30.000000 & 2880.01 & 1 & 1600 & 5788 & 400 & 4 & 1 & 1.27006 & 1.29543 & 8.896 \\
30 & 29.999920 & 2879.98 & 2 & 1160 & 3951 & 290 & 4 & 1 & 1.11196 & 1.13188 & 7.474 \\
30 & 29.999872 & 2879.98 & 3 & 896 & 3658 & 224 & 4 & 1 & 1.08456 & 1.10425 & 5.359 \\
30 & 30.000152 & 2880.01 & 4 & 812 & 2896 & 203 & 4 & 1 & 1.05696 & 1.07627 & 6.376 \\
30 & 30.000240 & 2880.03 & 5 & 684 & 2939 & 171 & 4 & 1 & 1.04522 & 1.06427 & 6.427 \\
30 & 29.999424 & 2879.96 & 6 & 672 & 2310 & 168 & 4 & 1 & 1.03745 & 1.05605 & 5.727 \\
30 & 29.999920 & 2879.98 & 8 & 536 & 2376 & 134 & 4 & 1 & 1.02961 & 1.04801 & 5.891 \\
30 & 29.998800 & 2879.86 & 10 & 520 & 1804 & 130 & 4 & 1 & 1.02253 & 1.04093 & 4.994 \\
30 & 30.000376 & 2880.03 & 12 & 476 & 1640 & 119 & 4 & 1 & 1.02062 & 1.03925 & 3.781 \\
30 & 29.999480 & 2879.96 & 16 & 380 & 1681 & 95 & 4 & 1 & 1.01724 & 1.03561 & 4.220 \\
30 & 30.000240 & 2880.03 & 20 & 344 & 1471 & 86 & 4 & 1 & 1.01641 & 1.03463 & 3.159 \\
30 & 30.000456 & 2880.03 & 24 & 324 & 1263 & 81 & 4 & 4 & 1.01553 & 1.03387 & 4.044 \\
30 & 29.999384 & 2879.96 & 32 & 268 & 1199 & 67 & 4 & 4 & 1.01257 & 1.03105 & 4.254 \\
30 & 30.002400 & 2880.24 & 40 & 240 & 1071 & 60 & 4 & 1 & 1.01465 & 1.03312 & 4.168 \\
30 & 30.000848 & 2880.08 & 48 & 232 & 873 & 58 & 4 & 4 & 1.01426 & 1.03281 & 4.178 \\
30 & 30.003096 & 2880.29 & 64 & 204 & 733 & 51 & 4 & 1 & 1.01638 & 1.03524 & 4.369 \\
50 & 50.000000 & 4800.00 & 1 & 2000 & 8113 & 500 & 4 & 1 & 1.24109 & 1.26573 & 8.740 \\
50 & 49.999992 & 4800.00 & 2 & 1428 & 5703 & 357 & 4 & 1 & 1.06784 & 1.08689 & 7.486 \\
50 & 49.999864 & 4799.97 & 3 & 1172 & 4636 & 293 & 4 & 1 & 1.03987 & 1.05843 & 5.577 \\
50 & 50.000200 & 4800.02 & 4 & 1060 & 3678 & 265 & 4 & 1 & 1.01215 & 1.03089 & 4.562 \\
50 & 50.000240 & 4800.02 & 5 & 904 & 3644 & 226 & 4 & 1 & 1.00187 & 1.02021 & 5.466 \\
50 & 49.999608 & 4799.97 & 6 & 828 & 3310 & 207 & 4 & 1 & 0.99398 & 1.01202 & 5.637 \\
50 & 49.999992 & 4800.00 & 8 & 732 & 2755 & 183 & 4 & 1 & 0.98391 & 1.00175 & 4.758 \\
50 & 49.999800 & 4799.97 & 10 & 668 & 2366 & 167 & 4 & 1 & 0.97787 & 0.99569 & 4.448 \\
50 & 49.999992 & 4800.00 & 12 & 588 & 2333 & 147 & 4 & 1 & 0.97464 & 0.99238 & 4.067 \\
50 & 50.000552 & 4800.05 & 16 & 532 & 1847 & 133 & 4 & 1 & 0.97092 & 0.98895 & 3.293 \\
50 & 50.000400 & 4800.02 & 20 & 456 & 1808 & 114 & 4 & 1 & 0.96882 & 0.98700 & 3.637 \\
50 & 50.000704 & 4800.07 & 24 & 416 & 1654 & 104 & 4 & 4 & 0.96618 & 0.98406 & 3.475 \\
50 & 49.999024 & 4799.91 & 32 & 344 & 1569 & 86 & 4 & 1 & 0.96708 & 0.98499 & 3.580 \\
50 & 50.002520 & 4800.23 & 40 & 332 & 1207 & 83 & 4 & 2 & 0.96588 & 0.98393 & 3.133 \\
50 & 50.000104 & 4800.00 & 48 & 308 & 1065 & 77 & 4 & 2 & 0.96590 & 0.98410 & 2.722 \\
50 & 49.999320 & 4799.92 & 60 & 252 & 1141 & 63 & 4 & 1 & 0.96864 & 0.98698 & 3.043 \\
50 & 49.997304 & 4799.74 & 64 & 252 & 1038 & 63 & 4 & 1 & 0.96792 & 0.98595 & 3.853 \\
\end{longtable}
\noindent\textit{Checkpoint metadata.} The 15M rows use seeds 1001 and 1002, $T=1{,}440{,}006{,}144$ bytes, and the final checkpoint after 87,891 steps. Their training pool contains 900M unique bytes; selection-validation CE uses 4,096 document-contained windows disjoint from the probe and test data. Data-order seed 271828 is shared across both initializations. These rows enter the same configuration-mean tables, frontier and allocation fit as the other budgets.
\endgroup

\section{Within-budget association}\label{app:correlations}
Pearson correlations use configuration means, not individual checkpoints with unequal replication. Spearman correlations use the same rows and are unchanged by taking the logarithm. Table~\ref{tab:within} retains every budget, including weaker associations.  Correlations and minima summarize the finite candidate set of each budget.
\begin{table}[htbp]\centering\small
\begin{tabular}{rrrrrrr}\toprule
$P$ (M) & Shapes & Raw $r$ & Log $r$ & $\rho$ & Val-best $d$ & Min-$F$ $d$ \\ \midrule
0.3 & 12 & 0.707 & 0.646 & 0.657 & 6 & 1 \\
0.6 & 12 & 0.266 & 0.255 & 0.378 & 3 & 16 \\
1 & 12 & -0.188 & -0.292 & -0.049 & 5 & 1 \\
1.6 & 12 & -0.056 & -0.172 & -0.371 & 5 & 1 \\
2.3 & 12 & 0.176 & 0.116 & -0.168 & 6 & 20 \\
3 & 12 & 0.553 & 0.468 & 0.189 & 10 & 24 \\
5 & 12 & 0.612 & 0.541 & 0.483 & 8 & 24 \\
10 & 12 & 0.951 & 0.823 & 0.951 & 20 & 20 \\
15 & 7 & 0.950 & 0.864 & 0.929 & 24 & 24 \\
20 & 16 & 0.951 & 0.783 & 0.835 & 24 & 12 \\
30 & 16 & 0.963 & 0.847 & 0.812 & 32 & 20 \\
50 & 17 & 0.967 & 0.885 & 0.900 & 40 & 48 \\
\bottomrule\end{tabular}

\caption{All twelve within-budget comparisons. Val-best depth minimizes seed-mean selection-validation CE in the candidate set of that budget.}\label{tab:within}
\end{table}
\begin{figure}[p]\centering\includegraphics[width=.98\linewidth]{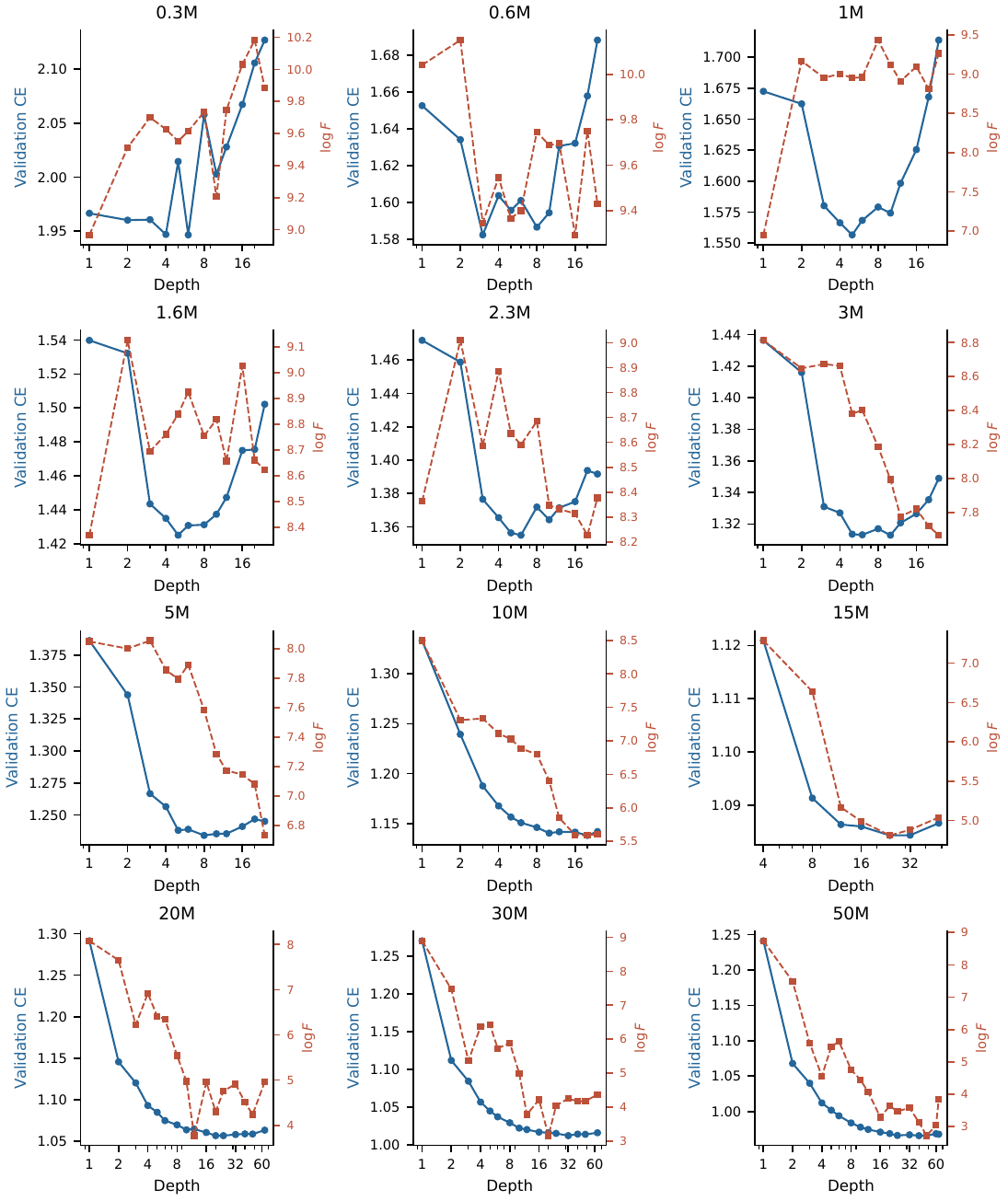}
\caption{All twelve budget scans. Blue axes show selection-validation CE; red axes show natural-log mean AGOP norm. Each panel has its own ranges. Curves join measured candidates only; the two vertical scales do not make their amplitudes directly comparable.}\label{fig:allbudgets}
\end{figure}
\FloatBarrier

\section{Frozen-checkpoint perturbation audit}\label{app:perturbation}
These measurements test the local input sensitivity of seven frozen checkpoints from the 15M budget. They are not used to choose frontier points or to validate the allocation fit. All are seed-1001 final checkpoints after 87,891 training steps. The Jacobian and Hessian identities quantify their local response under the recorded input coordinates.

\begin{figure}[htbp]\centering\includegraphics[width=.88\linewidth]{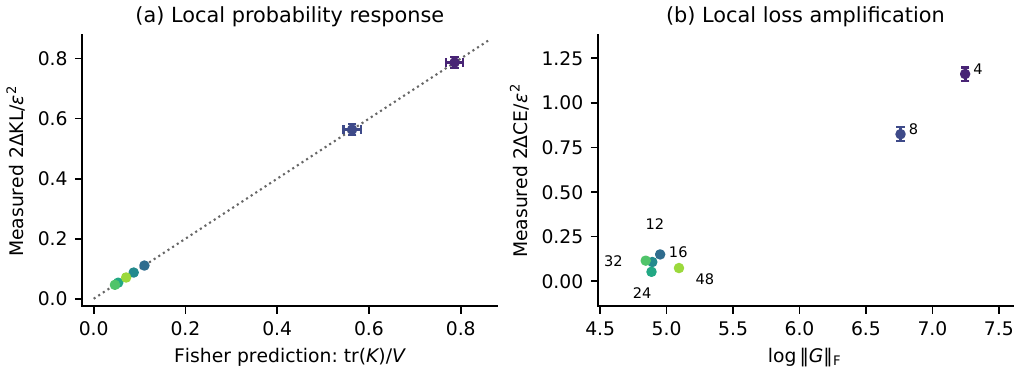}
\caption{Local response on the seven seed-1001 checkpoints from the 15M budget. Left: predictive KL curvature versus its Fisher/JVP prediction. Right: labeled CE curvature versus $\log F$, with depth labels. Error bars are direction SEM conditional on 96 examples and one trained checkpoint per shape. The reported $r(F,\mathrm{CE\ curvature})=0.9929$ uses $F$ before taking logs; it describes the local responses of these fixed functions.}\label{fig:bridge}
\end{figure}

The audit was specified before its execution but after the checkpoint losses and norms were known; it is a post hoc measurement check. All seven candidates, three radii, and 64 directions are retained. The same 96 document-contained AGOP examples (sampling seed 42) and Gaussian directions (seed 20260919) are shared across shapes. With $z\sim\mathcal N(0,I_{CV}/V)$, perturb embeddings by $\pm\epsilon zE$, leaving position embeddings unchanged. Model parameters and RMS normalizers are fixed. Both signs are evaluated to cancel odd-order terms.

For each direction compute $A(z)=M^{-1}\sum_i\|J_{f,i}z\|^2$ and $K(z)=M^{-1}\sum_i z^\top J_{\ell,i}^\top H_{\pi_i}J_{\ell,i}z$. Compare $A$ with the mean squared normalized-logit displacement divided by $\epsilon^2$, and $K$ with twice the symmetric predictive KL divided by $\epsilon^2$. The symmetric CE difference uses true held-out next-byte labels. At $\epsilon=0.01$, maximum relative discrepancies are \BridgeOutputError\% for output energy and \BridgeKLError\% for KL. Across the three radii, each shape's mean CE curvature varies by less than 0.060\%, supporting the local interpretation at these scales.

The initial float32 experiment with float64 reductions exhibited cancellation in small labeled-CE second differences. We repeated every candidate, direction, and radius in float64; both versions are retained in the evidence package. TF32 and mixed precision are disabled. Table~\ref{tab:bridgeradii} reports the float64 results. SEM and bootstrap intervals describe Monte Carlo directions conditional on fixed examples and checkpoints; they do not quantify population or training uncertainty. The experiment used four otherwise idle A100 GPUs and required about 39 GPU-minutes for the final double-precision evaluation.

\begin{table}[htbp]\centering\scriptsize
\begin{tabular}{rrrrrrr}\toprule
$d$ & $F$ & $\tau$ & $\tr K/V$ & $2\Delta\mathrm{CE}/\epsilon^2$ & $\widehat\Gamma/V$ & $\omega$ \\ \midrule
4 & 1402.75 & 8252.57 & 0.7866 & $1.1605\pm0.0365$ & $0.3739\pm0.0335$ & 0.00074 \\
8 & 862.85 & 5229.63 & 0.5632 & $0.8237\pm0.0381$ & $0.2605\pm0.0342$ & 0.00099 \\
12 & 141.36 & 1147.98 & 0.1106 & $0.1506\pm0.0077$ & $0.0400\pm0.0071$ & 0.00087 \\
16 & 133.26 & 779.01 & 0.0872 & $0.1080\pm0.0107$ & $0.0208\pm0.0095$ & 0.00099 \\
24 & 132.45 & 552.21 & 0.0534 & $0.0532\pm0.0085$ & $-0.0002\pm0.0083$ & 0.00080 \\
32 & 126.82 & 526.65 & 0.0463 & $0.1157\pm0.0068$ & $0.0694\pm0.0062$ & 0.00081 \\
48 & 162.87 & 566.40 & 0.0707 & $0.0742\pm0.0086$ & $0.0035\pm0.0076$ & 0.00105 \\
\bottomrule\end{tabular}

\caption{Local response at $\epsilon=0.01$, means with direction SEM. The residual estimate $\widehat\Gamma/V=2\Delta\mathrm{CE}/\epsilon^2-\E_zK(z)$ includes finite-radius error and does not independently verify the logit-Hessian term. An independent HVP check appears in Appendix~\ref{app:newaudit}. $\omega$ uses the shared-direction trace estimate.}\label{tab:bridgeresponse}
\end{table}
The positive second-order curvature contribution is substantial in several models. Depth 32, for example, has lower $F$ than depth 24 but a larger local CE response. This follows directly from the additional terms in Equation~\eqref{eq:lm_bridge}; an unweighted gradient norm need not preserve every loss ranking. The full-cohort Pearson correlations with CE curvature are high, while the corresponding Spearman coefficient for $F$ is 0.6429. The scalar norm and local CE curvature therefore resolve different aspects of input response. Trace and raw-logit comparisons in Table~\ref{tab:bridgecompare} show that the present evidence does not establish Frobenius norm as a uniquely superior diagnostic.
\begin{table}[htbp]\centering\small
\begin{tabular}{lrr}\toprule
Observable & $r$ with KL curvature & $r$ with CE curvature \\ \midrule
$F$ (RMS logits) & 0.9932 & 0.9929 \\
$\tau$ (RMS logits) & 0.9971 & 0.9966 \\
$F$ (raw logits) & 0.9845 & 0.9843 \\
$\tau$ (raw logits) & 0.9899 & 0.9896 \\
$\log F$ & 0.9931 & 0.9903 \\
Clean test CE & 0.8636 & 0.8631 \\
\bottomrule\end{tabular}

\caption{Comparator correlations over all seven shapes, with response measured at $\epsilon=0.01$. Raw-logit values restore the archived RMS factor; all rows use identical checkpoints.}\label{tab:bridgecompare}
\end{table}
\begin{table}[htbp]\centering\small
\begin{tabular}{rrrrr}\toprule
$d$ & Archived $F$ & Gaussian $F$ [95\% interval] & Archived $\tau$ & Gaussian $\tau$ \\ \midrule
4 & 1402.75 & 1334.23 [1092.59, 1547.06] & 8252.57 & 8790.20 \\
8 & 862.85 & 902.71 [738.76, 1044.84] & 5229.63 & 5343.99 \\
12 & 141.36 & 132.85 [103.07, 160.33] & 1147.98 & 1162.87 \\
16 & 133.26 & 142.58 [105.56, 175.41] & 779.01 & 785.03 \\
24 & 132.45 & 160.25 [110.59, 206.90] & 552.21 & 574.60 \\
32 & 126.82 & 144.97 [96.25, 187.85] & 526.65 & 529.81 \\
48 & 162.87 & 196.94 [136.92, 247.90] & 566.40 & 615.01 \\
\bottomrule\end{tabular}

\caption{Independent directional check of the operator. Gaussian estimates are $V\overline A$ for trace and $V\sqrt{\widehat{\operatorname{Var}}(A)/2}$ for norm. Brackets give a 5,000-resample, direction-bootstrap percentile interval, conditional on the examples. Archived estimates have their own finite-probe error.}\label{tab:bridgeoperator}
\end{table}
\begin{table}[htbp]\centering\small
\begin{tabular}{rrrrrr}\toprule
$d$ & $\epsilon$ & $A$ & Output curvature & KL curvature & CE curvature \\ \midrule
4 & 0.0025 & 34.33671 & 34.33755 & 0.78666 & $1.15993\pm0.03652$ \\
4 & 0.005 & 34.33671 & 34.34009 & 0.78675 & $1.16003\pm0.03652$ \\
4 & 0.01 & 34.33671 & 34.35023 & 0.78709 & $1.16051\pm0.03650$ \\
8 & 0.0025 & 20.87495 & 20.87551 & 0.56322 & $0.82336\pm0.03810$ \\
8 & 0.005 & 20.87495 & 20.87720 & 0.56329 & $0.82343\pm0.03810$ \\
8 & 0.01 & 20.87495 & 20.88398 & 0.56355 & $0.82373\pm0.03810$ \\
12 & 0.0025 & 4.54245 & 4.54250 & 0.11055 & $0.15053\pm0.00772$ \\
12 & 0.005 & 4.54245 & 4.54267 & 0.11056 & $0.15053\pm0.00772$ \\
12 & 0.01 & 4.54245 & 4.54333 & 0.11057 & $0.15055\pm0.00772$ \\
16 & 0.0025 & 3.06653 & 3.06658 & 0.08721 & $0.10797\pm0.01074$ \\
16 & 0.005 & 3.06653 & 3.06674 & 0.08722 & $0.10798\pm0.01074$ \\
16 & 0.01 & 3.06653 & 3.06735 & 0.08725 & $0.10802\pm0.01074$ \\
24 & 0.0025 & 2.24453 & 2.24460 & 0.05345 & $0.05318\pm0.00848$ \\
24 & 0.005 & 2.24453 & 2.24483 & 0.05346 & $0.05319\pm0.00848$ \\
24 & 0.01 & 2.24453 & 2.24574 & 0.05348 & $0.05321\pm0.00848$ \\
32 & 0.0025 & 2.06957 & 2.06962 & 0.04632 & $0.11571\pm0.00677$ \\
32 & 0.005 & 2.06957 & 2.06977 & 0.04632 & $0.11571\pm0.00677$ \\
32 & 0.01 & 2.06957 & 2.07036 & 0.04634 & $0.11573\pm0.00677$ \\
48 & 0.0025 & 2.40237 & 2.40242 & 0.07074 & $0.07421\pm0.00865$ \\
48 & 0.005 & 2.40237 & 2.40256 & 0.07075 & $0.07422\pm0.00865$ \\
48 & 0.01 & 2.40237 & 2.40314 & 0.07077 & $0.07424\pm0.00865$ \\
\bottomrule\end{tabular}

\caption{Every radius and shape in the float64 audit. Output curvature is symmetric normalized-logit squared change divided by $\epsilon^2$; KL and CE curvatures use twice the corresponding increment divided by $\epsilon^2$. $A$ is the JVP prediction.}\label{tab:bridgeradii}
\end{table}
\FloatBarrier

\section{Descriptive shape fitting}\label{app:fits}
\subsection{The validation-selected loss--AGOP frontier}
For each of the twelve budgets, including 15M, select the lowest seed-mean selection-validation CE before fitting test-loss associations. Equation~\eqref{eq:frontierfit} uses these twelve points. Equation~\eqref{eq:frontierfit} fits $b+c(F/1000)^\beta$ with $0\leq b<\min_iL_i$, $c\geq0$, $0.01\leq\beta\leq3$, and four deterministic exponent starts. The best least-squares objective is retained. Its offset reaches the upper bound, preventing interpretation as an identified asymptotic loss. The fit has $R^2=0.9448$, in-sample RMSE 0.06620, and leave-one-budget-out RMSE 0.10833. No candidate selection or training uses this test curve.

The corresponding validation fit has $R^2=0.9374$. A budget-only validation fit $E+Ap^{-\alpha}$ gives $E=0.7741$, $A=0.7588$, $\alpha=0.3319$, $R^2=0.9877$, and leave-one-budget-out RMSE 0.06405. Budget alone therefore predicts frontier validation loss more accurately in this scan. Exposure and budget are not independently varied, so these fits describe the recorded parameter--data path.

\subsection{The full shape landscape}
The budget-dependent quadratic model used in the main text is
\begin{equation}\label{eq:shape}
 \widehat L(P,d)=E+A p^{-\alpha}+B p^{-\gamma}(\log d-\mu-\eta\log p)^2,
 \qquad p=P_0/10^6.
\end{equation}
Here $P_0$ is nominal parameter budget.
We fit validation CE with three models: $E+Ap^{-\alpha}$; $E+Ad^{-\alpha}+B(w/128)^{-\beta}$; and Equation~\eqref{eq:shape}. Each budget receives total weight one, divided equally among its means. Bounded least squares uses eight deterministic starts (seed 883), at most 2,500 evaluations per start, and the lowest converged objective. Every model/fold has a converged start. The depth--width fit reaches the lower bound $E=0$, limiting coefficient identification.

Each fold trains on eleven budgets and evaluates the twelfth. This is ordinary leave-one-budget-out cross-validation, with 15M treated in the same way as every other left-out budget. The main coefficients in Table~\ref{tab:fitcoeffs} use all twelve budgets. A selection gap is the held-out validation loss of the predicted-best scanned candidate minus that budget's observed minimum. It does not use test CE to select a candidate. The budget-only model ties all shapes, so no selection gap is assigned to it. Fold MSE averaged before taking the square root gives RMSE 0.08436, 0.07198 and 0.07009 for budget, depth--width and shape, respectively. Mean validation gaps for the last two are 0.00159 and 0.00605; signed test gaps are 0.00146 and 0.00588.

\begin{table}[htbp]\centering\small
\begin{tabular}{lrrrrrrrr}\toprule
Model & $E$ & $A$ & $\alpha$ & $B$ & $\beta$ & $\mu$ & $\eta$ & $\gamma$ \\ \midrule
budget & 0.8277 & 0.7469 & 0.3561 & $-$ & $-$ & $-$ & $-$ & $-$ \\
depth--width & 9.148e-26 & 1.045 & 0.2653 & 0.8348 & 0.4859 & $-$ & $-$ & $-$ \\
shape & 0.7223 & 0.8193 & 0.3087 & 0.03659 & $-$ & 1.667 & 0.459 & 0.1665 \\
\bottomrule\end{tabular}

\caption{Descriptive fit coefficients using all 152 means at twelve budgets. Near-zero $E$ is an optimizer boundary, not an identified zero irreducible loss.}\label{tab:fitcoeffs}
\end{table}
\begin{table}[htbp]\centering\small
\begin{tabular}{rrrrrr}\toprule
$P$ (M) & Budget RMSE & DW RMSE & Shape RMSE & DW gap & Shape gap \\ \midrule
0.3 & 0.19603 & 0.19037 & 0.18729 & 0.01413 & 0.01413 \\
0.6 & 0.13651 & 0.12864 & 0.13413 & 0.00000 & 0.02135 \\
1 & 0.07036 & 0.05014 & 0.05603 & 0.00000 & 0.00000 \\
1.6 & 0.03932 & 0.02798 & 0.01787 & 0.00000 & 0.00572 \\
2.3 & 0.03708 & 0.03603 & 0.01588 & 0.00000 & 0.01688 \\
3 & 0.04080 & 0.03143 & 0.01606 & 0.00000 & 0.00407 \\
5 & 0.05020 & 0.02235 & 0.01721 & 0.00000 & 0.00132 \\
10 & 0.05892 & 0.01667 & 0.01330 & 0.00000 & 0.00330 \\
15 & 0.02732 & 0.01333 & 0.00766 & 0.00203 & 0.00168 \\
20 & 0.05792 & 0.02421 & 0.01427 & 0.00002 & 0.00002 \\
30 & 0.06334 & 0.02548 & 0.01806 & 0.00296 & 0.00296 \\
50 & 0.06856 & 0.04013 & 0.02915 & 0.00000 & 0.00119 \\
\bottomrule\end{tabular}

\caption{Each row is a held-out budget in leave-one-budget-out cross-validation. DW is the separable depth--width model; Shape is Equation~\eqref{eq:shape}. Gaps are selection-validation CE above the best observed candidate.}\label{tab:fitcv}
\end{table}

\section{Additional measurement controls}\label{app:newaudit}
\subsection{Matched constant baselines and toy decomposition}\label{app:toycontrols}
We replay the archived 22-size, five-seed protocol, without changing its optimizer, 3,000-step horizon, initialization seeds or data-generation seeds. The CUDA generator uses training seed $12345+n$ and test seed $54321+n$ with the original generation code. Every test-loss value matches the archive to $2.3\times10^{-16}$ in the stored scalars. Dataset checksums, replayed weights and per-run diagnostics are retained. This is a fixed-protocol reconstruction for measurement, not a training-duration comparison.

For each actual training set, let $\widehat\mu=n^{-1}\sum_i x_i$. Because coordinates are nonnegative, $f(x)=\widehat\mu$ is attainable by $W=0$, $b=\widehat\mu$. It minimizes empirical squared loss over all constant predictors, including the bias-only ReLU family. We solve this baseline analytically rather than adding a separate optimization error. Training-mean and bias-only ERM predictions therefore coincide, with $F=0$, and are evaluated on exactly the same generated test data as the full network. This differs from using an oracle population mean. The population identity is
\begin{equation*}
 L_{\rm const}=\E\|x\|^2-\|\mu\|^2,\qquad
 \E_{\rm train}L(\widehat\mu)=L_{\rm const}(1+1/n),
\end{equation*}
for iid training draws and an independent population test, with $\mu=\E x$. At $n=40{,}000$, the measured mean/bias-only loss is $0.992437$, below the full network's $0.993247$. All configuration means from $n=1946$ onward are above the matched constant baseline. This limits an interpretation of the later descent as evidence of useful input-dependent features.

We recompute $q_{\rm gate},r_{\rm eff},B,C_x$ from each replayed checkpoint in float64. Each decomposition uses its actual test-set energy $v$; it closes within $2.45\times10^{-15}$. These diagnostics recompute gate decisions and norm at higher precision; their $F$ values differ from the archived reconstructed norms by at most 0.245\% per run. Figure~\ref{fig:dd} and its full-sweep fit retain the archived norms. In Table~\ref{tab:toycontrols}, products are evaluated per run before averaging, so products of displayed means need not reproduce a mean loss. These quantities check a sufficient explanation of the observed association; they do not treat the norm as an independently controlled mechanism. The affine-fit residual is $L-(L_0+aF)$ from Equation~\eqref{eq:dd_affine}, not a nonnegative excess-loss component.
\begin{table}[htbp]\centering\scriptsize\setlength{\tabcolsep}{2.2pt}
\begin{tabular}{rrrrrrrrrr}\toprule
$n$ & Full test & Mean/bias & $q_{\rm gate}$ & $r_{\rm eff}$ & $a$ & $B$ & $C_x$ & $h$ & Affine residual \\ \midrule
3 & 1.102627 & 1.318494 & 0.0011 & 1.889 & 0.0015 & 0.1086 & 0.0050 & 0.0986 & 0.0988 \\
5 & 0.998945 & 1.195262 & 0.0123 & 1.335 & 0.0146 & -0.0015 & 0.0009 & -0.0034 & -0.0032 \\
8 & 1.007707 & 1.117356 & 0.0356 & 1.174 & 0.0400 & -0.0046 & 0.0006 & -0.0057 & 0.0058 \\
10 & 1.007314 & 1.085834 & 0.0085 & 1.618 & 0.0107 & 0.0029 & 0.0007 & 0.0015 & 0.0052 \\
15 & 1.003641 & 1.056123 & 0.0156 & 1.885 & 0.0215 & -0.0039 & 0.0005 & -0.0048 & 0.0016 \\
30 & 1.010302 & 1.025504 & 0.0340 & 1.636 & 0.0434 & -0.2408 & 0.0018 & -0.2443 & 0.0042 \\
50 & 1.009268 & 1.011357 & 0.0258 & 1.752 & 0.0340 & -0.2212 & 0.0015 & -0.2241 & 0.0025 \\
100 & 1.032078 & 1.002292 & 0.0150 & 1.800 & 0.0200 & -0.7951 & 0.0034 & -0.8018 & -0.0016 \\
200 & 1.248293 & 0.997102 & 0.0086 & 1.887 & 0.0118 & -4.1233 & 0.0164 & -4.1561 & -0.0422 \\
500 & 1.534006 & 0.994696 & 0.0064 & 1.821 & 0.0086 & -5.1217 & 0.0306 & -5.1829 & 0.0227 \\
1000 & 1.058263 & 0.993599 & 0.0073 & 1.799 & 0.0097 & -0.5010 & 0.0090 & -0.5189 & 0.0150 \\
1395 & 1.000026 & 0.993374 & 0.0044 & 1.857 & 0.0059 & -0.0429 & 0.0057 & -0.0543 & -0.0081 \\
1946 & 0.994491 & 0.993203 & 0.0031 & 1.991 & 0.0043 & -0.0035 & 0.0072 & -0.0180 & -0.0094 \\
2714 & 0.993201 & 0.992743 & 0.0033 & 1.994 & 0.0046 & -0.0019 & 0.0075 & -0.0168 & -0.0102 \\
3786 & 0.993174 & 0.992628 & 0.0026 & 1.997 & 0.0037 & -0.0005 & 0.0070 & -0.0145 & -0.0102 \\
5282 & 0.992724 & 0.992556 & 0.0029 & 1.996 & 0.0041 & -0.0009 & 0.0074 & -0.0157 & -0.0106 \\
7368 & 0.993290 & 0.992770 & 0.0035 & 1.994 & 0.0050 & -0.0021 & 0.0070 & -0.0162 & -0.0099 \\
10278 & 0.992984 & 0.992696 & 0.0032 & 1.995 & 0.0045 & -0.0015 & 0.0072 & -0.0159 & -0.0103 \\
14337 & 0.993052 & 0.992467 & 0.0029 & 1.997 & 0.0041 & -0.0015 & 0.0068 & -0.0152 & -0.0102 \\
20000 & 0.993113 & 0.992660 & 0.0034 & 1.986 & 0.0048 & -0.0019 & 0.0070 & -0.0160 & -0.0101 \\
30000 & 0.993586 & 0.992685 & 0.0027 & 1.996 & 0.0038 & -0.0015 & 0.0064 & -0.0144 & -0.0097 \\
40000 & 0.993247 & 0.992437 & 0.0027 & 1.997 & 0.0038 & -0.0017 & 0.0066 & -0.0149 & -0.0101 \\
\bottomrule\end{tabular}

\caption{All toy sizes: matched baselines, measured decomposition terms and full-sweep affine residuals. Five-seed means; $a=q_{\rm gate}\sqrt{r_{\rm eff}}$, $h=B-2C_x$. Neither $a$ nor $h$ is approximately constant across the complete trajectory.}\label{tab:toycontrols}
\end{table}

\subsection{Coordinate, normalization and estimator checks}
These checkpoints are drawn from the 15M shapes already used as fitting points. The controls in this and the following subsection concern the AGOP estimator and local CE response, not shape validation.

All seven seed-1001 checkpoints at the 15M budget are remeasured with the archived 96 examples and shared 96 Rademacher probes. We compare ambient $G$ and simplex-tangent $P_TGP_T$, rejecting nonfinite derivatives. The tangent computation projects both input probe and operator product. Raw/normalized norms, trace, RMS, effective rank and last/all-position clean CE are retained. Operator products and probe CE use float32 model computation with float64 reductions; the curvature checks below use float64 throughout. Probe CE uses only these 96 examples, whereas full-test CE retains its archived evaluation set and precision. Tiny differences between their absolute values are not interpreted as architecture effects.

Depths 24 and 32 receive separate estimator-size checks: $(M,R)=(96,96),(96,192),(192,96)$, using nested example and probe prefixes. Each sample size has its measured RMS, fixed during differentiation; raw operator values are also reported to distinguish normalization effects. Probe standard errors are conditional on examples and checkpoints. The cross-example $U_z$ estimator removes empirical diagonal terms in expectation under iid sampling, but remains noisy and does not measure training-seed uncertainty. No main scan scalar is replaced by whichever estimator produces preferred agreement.

\subsection{Independent nonlinear-curvature measurements}
For every seed-1001 shape, and additionally seed 1002 at depths 24 and 32, we directly estimate the logit-Hessian contribution. With $r=\pi-e_y$ held constant at the unperturbed input, a Hessian-vector product of $r^\top\ell(q)$ gives
\begin{equation*}
 z^\top\nabla_q^2[r^\top\ell(q)]z
 =\sum_k r_k z^\top\nabla_q^2\ell_k(q)z.
\end{equation*}
Holding $r$ fixed is essential; differentiating it would mix the Fisher term into this quantity. Averaging over $z\sim\mathcal N(0,I/V)$ estimates $\Gamma/V$ independently of finite CE differences. We compare the sum of direct Fisher and HVP estimates with antithetic $2\Delta\mathrm{CE}/\epsilon^2$ at $\epsilon=0.005$, in float64, for both ambient and projected directions. There are 64 directions for depths 24 and 32 in each seed, and 16 for the other five checkpoints. All use the same 96-example prefix. Projected probes have covariance $P_T/V$ and therefore test the tangent-space analogue of Equation~\eqref{eq:lm_bridge}.

\paragraph{Coordinate and output comparisons.}
The largest tangent-versus-ambient norm change is 25.634\% at these checkpoints. The minimum occurs at depth 24 for the tangent measurement, versus 32 for the ambient measurement. This tests the chosen parameterization only: the comparison does not make ambient AGOP invariant to the embedding shift described in Appendix~\ref{app:estimator}. Table~\ref{tab:newoperators} also exposes changes in logit RMS and effective rank; $r_{\rm eff}$ spans 12.09--65.95. Correlations with position-aligned probe CE and full-test CE are reported separately, rather than treating the small probe sample as another full test.
\begin{table}[htbp]\centering\scriptsize\setlength{\tabcolsep}{3pt}
\begin{tabular}{rrrrrrrrrr}\toprule
$d$ & $F$ & $F_T$ & Raw $F$ & $\tau$ & $s_\ell$ & $r_{\rm eff}$ & $\omega$ & Last CE & All CE \\ \midrule
4 & 1402.75 & 1082.41 & 43546.1 & 8252.6 & 5.572 & 34.61 & 0.0008 & 0.6883 & 1.1257 \\
8 & 862.85 & 712.82 & 23624.6 & 5229.6 & 5.233 & 36.73 & 0.0010 & 0.6858 & 1.0923 \\
12 & 141.36 & 127.95 & 3943.3 & 1148.0 & 5.282 & 65.95 & 0.0009 & 0.6908 & 1.0908 \\
16 & 133.26 & 101.73 & 3840.1 & 779.0 & 5.368 & 34.17 & 0.0010 & 0.6855 & 1.0901 \\
24 & 132.45 & 98.50 & 3940.0 & 552.2 & 5.454 & 17.38 & 0.0008 & 0.6733 & 1.0884 \\
32 & 126.82 & 99.87 & 3517.6 & 526.6 & 5.267 & 17.25 & 0.0008 & 0.6880 & 1.0900 \\
48 & 162.87 & 125.92 & 4561.4 & 566.4 & 5.292 & 12.09 & 0.0011 & 0.6484 & 1.0942 \\
\bottomrule\end{tabular}

\caption{Seed-1001 coordinate and scale controls at all seven 15M shapes, with $M=R=96$. $F_T$ denotes the simplex-tangent norm; the raw-logit norm is $s_\ell^2F$. Clean CE uses the same 96 examples; $\omega$ uses the shared Gaussian directional Fisher/response estimates.}\label{tab:newoperators}
\end{table}
\begin{table}[htbp]\centering\small
\begin{tabular}{lrrr}\toprule
Diagnostic & Full test CE & Probe last CE & Probe all CE \\ \midrule
F & 0.9056 & 0.3013 & 0.8667 \\
tangent F & 0.8891 & 0.3111 & 0.8467 \\
Raw-logit norm & 0.9309 & 0.2981 & 0.8968 \\
trace & 0.8957 & 0.3529 & 0.8509 \\
Raw-logit trace & 0.9230 & 0.3463 & 0.8833 \\
logit RMS & 0.7078 & 0.0884 & 0.7297 \\
\bottomrule\end{tabular}

\caption{Pearson comparisons on the same seven seed-1001 shapes. Full test CE is the original all-position evaluation; last/all probe CE use only the 96 diagnostic examples. These correlations are descriptive, not independent generalization estimates.}\label{tab:newcomparators}
\end{table}
\paragraph{Estimator sensitivity.}
Table~\ref{tab:estimatorcontrols} separates more probes at fixed examples from more examples at fixed probes. Probe SE is a delta-method SE conditional on the chosen sample, not a data or training-seed interval. The diagonal correction $\widehat U$ quantifies why the empirical squared norm need not equal a population squared norm. At depths 24 and 32, doubling probes changes $F$ by less than 0.2\%, whereas doubling examples reduces it by about 20.6\% and 14.1\%, respectively. This is a finite-sample comparison, not a convergence guarantee. Main-scan measurements retain the original common estimator size.
\begin{table}[htbp]\centering\small
\begin{tabular}{rrrrrrrr}\toprule
$d$ & $M$ & $R$ & $\widehat F$ & Probe SE & Change (\%) & $\widehat U/\widehat Q$ & $\sqrt{\widehat U}/s_\ell^2$ \\ \midrule
24 & 96 & 96 & 132.452 & 7.168 & +0.00 & 0.712 & 111.758 \\
24 & 96 & 192 & 132.602 & 5.026 & +0.11 & 0.715 & 112.133 \\
24 & 192 & 96 & 105.215 & 5.079 & -20.56 & 0.805 & 94.426 \\
32 & 96 & 96 & 126.815 & 6.050 & +0.00 & 0.769 & 111.194 \\
32 & 96 & 192 & 126.605 & 4.351 & -0.17 & 0.775 & 111.423 \\
32 & 192 & 96 & 108.911 & 5.132 & -14.12 & 0.893 & 102.896 \\
\bottomrule\end{tabular}

\caption{Nested estimator checks. Changes are relative to $(96,96)$ within each depth; the output RMS is held fixed within each measurement.}\label{tab:estimatorcontrols}
\end{table}
\paragraph{Independent CE-curvature closure.}
Across the nine checkpoints and both coordinate choices, direct Fisher plus HVP agrees with the symmetric finite CE curvature to maximum absolute difference 9.41e-05 (maximum relative difference 0.016\%). This independently checks the nonlinear contribution rather than defining it by subtraction. For seed 1001 in ambient coordinates, depth 32 has smaller Fisher curvature than depth 24 (0.04632 versus 0.05345), but a larger HVP contribution (0.06938 versus $-0.00027$), yielding greater CE curvature (0.11571 versus 0.05319). For seed 1002 the respective CE curvatures are 0.14254 and 0.09804; the same depth ordering also holds for tangent directions in both seeds. These checks explain local response differences conditional on the checkpoints, without attributing their clean-test differences to input noise.
\begin{table}[htbp]\centering\scriptsize\setlength{\tabcolsep}{3pt}
\begin{tabular}{rrcrrrrrr}\toprule
$d$ & Seed & Coord. & $R$ & Fisher & HVP $\Gamma/V$ & CE curvature & Dir. SEM & Closure diff. \\ \midrule
4 & 1001 & A & 16 & 0.75864 & 0.41193 & 1.17064 & 0.06021 & 7.63e-05 \\
4 & 1001 & T & 16 & 0.64322 & 0.36700 & 1.01031 & 0.05036 & 9.02e-05 \\
8 & 1001 & A & 16 & 0.58369 & 0.29568 & 0.87946 & 0.09001 & 9.41e-05 \\
8 & 1001 & T & 16 & 0.50089 & 0.29185 & 0.79280 & 0.08748 & 6.62e-05 \\
12 & 1001 & A & 16 & 0.11499 & 0.04608 & 0.16108 & 0.01862 & 5.86e-06 \\
12 & 1001 & T & 16 & 0.10834 & 0.05728 & 0.16562 & 0.01420 & 2.29e-06 \\
16 & 1001 & A & 16 & 0.08701 & 0.01489 & 0.10191 & 0.01828 & 1.13e-05 \\
16 & 1001 & T & 16 & 0.07265 & 0.01125 & 0.08391 & 0.01613 & 7.63e-06 \\
24 & 1001 & A & 64 & 0.05345 & -0.00027 & 0.05319 & 0.00848 & 8.45e-06 \\
24 & 1001 & T & 64 & 0.04227 & -0.00581 & 0.03646 & 0.00693 & 5.65e-06 \\
32 & 1001 & A & 64 & 0.04632 & 0.06938 & 0.11571 & 0.00677 & 5.25e-06 \\
32 & 1001 & T & 64 & 0.03661 & 0.05408 & 0.09069 & 0.00547 & 4.69e-06 \\
48 & 1001 & A & 16 & 0.06724 & 0.01801 & 0.08526 & 0.01684 & 9.94e-06 \\
48 & 1001 & T & 16 & 0.05177 & 0.01710 & 0.06888 & 0.01086 & 9.56e-06 \\
24 & 1002 & A & 64 & 0.06185 & 0.03618 & 0.09804 & 0.00979 & 3.20e-06 \\
24 & 1002 & T & 64 & 0.05108 & 0.03388 & 0.08496 & 0.00654 & 2.71e-06 \\
32 & 1002 & A & 64 & 0.06878 & 0.07375 & 0.14254 & 0.01005 & 1.13e-05 \\
32 & 1002 & T & 64 & 0.05440 & 0.05490 & 0.10931 & 0.00756 & 9.15e-06 \\
\bottomrule\end{tabular}

\caption{Independent Hessian-vector checks in float64 at $\epsilon=0.005$. A/T denote ambient/tangent directions. Fisher plus HVP predicts CE curvature; the last column is measured minus predicted. SEM is over shared directions conditional on 96 examples. For T, $\Gamma/V$ denotes the corresponding projected trace.}\label{tab:hvpcontrols}
\end{table}

\FloatBarrier

\section{Extended related work}\label{app:related}
\subsection{Scaling laws}
The scaling literature progressed from reproducible empirical curves to prescriptions for allocating finite resources. \citet{hestness2017} measured power-law generalization trends in translation, language modeling, vision and speech. \citet{rosenfeld2020} developed joint model--data error predictions and tested extrapolation across scales. For language models, \citet{kaplan2020} organized loss around parameter count, data and compute, reporting comparatively weak shape dependence over much of their explored range. \citet{henighan2020} tested autoregressive scaling in additional domains; \citet{brown2020} demonstrated improved few-shot task performance with increasing language-model scale. These observations concern measured regimes and tasks, rather than a guarantee that every additional resource improves every capability.

Chinchilla showed why adding parameters alone can waste a training-compute budget: the allocation between model size and training tokens must be considered jointly \citep{hoffmann2022}. Subsequent studies made the resource definition more precise. Data pruning changes which examples are retained \citep{sorscher2022}, while data-constrained scaling distinguishes additional unique examples from repeated exposure \citep{muennighoff2025}. \citet{bahri2024} distinguish variance- and resolution-limited mechanisms, demonstrating that similar-looking power laws can arise for different reasons. Gemstones shows that architecture, learning rate, cooldown and checkpoint selection can affect a fitted prescription \citep{mcleish2025}. These findings make architecture and training part of the interpretation of a scaling law. Our question concerns allocation within a specified parameter budget and recorded exposure schedule; it is complementary to compute-optimal model--data allocation and does not establish matched-compute efficiency.

\subsection{Model shape}
Theoretical shape studies ask which functions benefit from depth. \citet{telgarsky2016} construct functions efficiently represented by deep networks that require exponentially larger shallow approximations. This expressive separation does not itself determine the best shape for language data or a finite training recipe. \citet{levine2020} specialize the allocation question to self-attention, predicting a width-dependent transition between depth efficiency and inefficiency and comparing depth--width allocations empirically. EfficientNet instead studies the empirical balance of depth, width and input resolution for convolutional vision models \citep{tan2019}; its compound-scaling principle illustrates coordinated architecture allocation without yielding a Transformer-specific rule.

Task-based comparisons ask whether depth changes what a model learns. \citet{petty2024} compare fixed-parameter Transformer families and evaluate both language modeling and compositional generalization after fine-tuning. They find diminishing depth returns, emphasizing that the task and latency objective matter. Gemstones examines a broader architecture and training landscape \citep{mcleish2025}. CompleteP explains how depth-aware hyperparameter transfer and non-lazy feature learning can make additional shapes compute-efficient \citep{dey2025}. \citet{liao2026} replace conventional FFNs with hourglass structures and decouple attention width from residual width, measuring language performance and systems efficiency. Their results show that a depth--width comparison can change internal bottlenecks and optimization behavior as well as layer count. Our matched-parameter scan likewise evaluates complete configurations, not an isolated causal effect of depth.

Representation and processing arguments offer a mechanistic motivation for these comparisons. \citet{liu2025} derive approximately inverse-width representation-error scaling in a strong-superposition regime. Their discussion explicitly distinguishes representation from processing loss and proposes balancing them under a size constraint; it would therefore be inaccurate to say the paper ignores depth altogether. The inverse-width result alone does not identify the processing term or the best trained shape. More recent work connects inverse-depth behavior to ensemble-like averaging across similar layers \citep{liu2026depth} and argues that scaling coefficients determine preferred shapes within a proposed universality class \citep{liu2026coefficients}. We investigate the functional response geometry accompanying favorable allocations, rather than separately identifying those latent loss components.

Our looped-Transformer discussion suggests a future comparison across allocation mechanisms. Universal Transformers reuse parameters along recurrent depth \citep{dehghani2019}; recurrent-depth language models also spend additional latent computation at inference \citep{geiping2025}. Recurrence and an additional unique block can have different scaling benefits \citep{schwethelm2026}. A controlled comparison should therefore specify stored parameter count, training exposure and inference compute separately. It could test whether useful sequential computation reduces error-bearing responses until representation or sharing constraints dominate, or whether repeated Jacobian products amplify unstable directions. Our untied-model scan does not choose between those outcomes.

\subsection{Superposition hypothesis}
Polysemanticity concerns neurons responding to several unrelated concepts, which makes neuron-level explanations incomplete \citep{olah2020}. Superposition offers a representational explanation: sparse features can share nonorthogonal directions when their number exceeds the available dimensions. The tied toy autoencoder makes this packing and its interference accessible through a weight Gram matrix \citep{elhage2022}. Its assumptions are useful for isolating geometry, but the resulting single-bottleneck calculation is not an end-to-end description of a general deep computation.

SAEs address the interpretation problem by learning a sparse dictionary of activation directions. \citet{bricken2023} demonstrate dictionary learning in a one-layer language model, and \citet{huben2024} evaluate the interpretability and behavioral relevance of recovered features. TopK SAEs directly control the number of active latents \citep{gao2025}; JumpReLU improves reconstruction at a chosen sparsity level \citep{rajamanoharan2024}. These are approaches to decomposing learned activations, not guarantees that the original network becomes free of superposition. Many implementations target one layer, but this is not a universal limitation: sparse crosscoders jointly read and reconstruct multiple layers or models \citep{lindsey2024}. Cross-layer feature recovery and functional sensitivity remain distinct measurements.

NFA supplies a route from weight geometry to gradients. \citet{radhakrishnan2023} connect feature learning to layerwise AGOPs; \citet{beaglehole2024} analyze the development of weight--gradient alignment. \citet{tansley2025} prove depth-dependent matrix-power relations in specified linear settings and exhibit nonlinear exceptions even with arbitrarily good data fit. Thus NFA motivates an observable without supplying a universal proportionality theorem for the Transformers in our scan. Earlier Jacobian work already links sensitivity to generalization empirically and through margin-based guarantees \citep{novak2018,sokolic2017}. Our focus is the Frobenius norm of the averaged input operator, its controlled shared-amplification mechanism, and its association with architecture. The norm reflects response strength and overlap; semantic feature multiplicity is not identified by that scalar alone.

\end{document}